\documentclass[final,12pt]{colt2023} % Anonymized submission
    \PassOptionsToPackage{numbers, compress}{natbib}
\usepackage{amsmath}
\usepackage{pifont}
\usepackage[shortlabels]{enumitem}

\newcommand{\cmark}{{\color{olive} \text{\ding{51}}}}
\newcommand{\xmark}{{\color{red} \text{\ding{56}}}}

\usepackage[utf8]{inputenc} % allow utf-8 input
\usepackage[T1]{fontenc}    % use 8-bit T1 fonts
\usepackage{hyperref}       % hyperlinks
\usepackage{booktabs}       % professional-quality tables
\usepackage{amsfonts}       % blackboard math symbols
\usepackage{nicefrac}       % compact symbols for 1/2, etc.
\usepackage{microtype}      % microtypography
\usepackage{xcolor}         % colors

\usepackage{framed}
\definecolor{light-gray}{gray}{0.96}
\colorlet{shadecolor}{light-gray}

\usepackage[font=small,labelfont=bf,tableposition=top]{caption}
\usepackage{amsmath,amsfonts,bm}

\def\figref#1{figure~\ref{#1}}
\def\Figref#1{Figure~\ref{#1}}

\def\eqref#1{equation~\ref{#1}}
\providecommand{\defref}[1]{definition~\ref{#1}}

\providecommand{\thmref}[1]{theorem~\ref{#1}}

\providecommand{\lemref}[1]{lemma~\ref{#1}}

\def\1{\bm{1}}

\def\eps{{\epsilon}}

\def\rw{{\textnormal{w}}}
\def\rx{{\textnormal{x}}}
\def\ry{{\textnormal{y}}}

\DeclareMathAlphabet{\mathsfit}{\encodingdefault}{\sfdefault}{m}{sl}
\SetMathAlphabet{\mathsfit}{bold}{\encodingdefault}{\sfdefault}{bx}{n}

\DeclareMathOperator*{\argmin}{arg\,min}

\newcommand{\E}[1]{\mathbb{E} \left[ #1 \right]}

\providecommand{\inner}[2]{ \langle #1, #2 \rangle}

\providecommand{\norm}[1]{\ensuremath{\left\| #1 \right\|}}

\makeatletter
\newcommand{\biggg}{\bBigg@{3}}
\newcommand{\vast}{\bBigg@{4}}
\newcommand{\Vast}{\bBigg@{5}}
\makeatother

\renewcommand{\varrho}{\rho}

\usepackage{mathtools}
\DeclareMathOperator{\R}{ReLU}
\DeclareMathOperator{\vol}{vol}
 \newtheorem*{rem}{Remark}
  \newtheorem*{exm}{Example}

 \newcommand{\tup}{^{(t+1)}}

 \newcommand{\tu}{^{(t)}}

\newcommand{\RR}{\mathbb{R}}
\DeclareMathOperator{\sign}{sign}

\providecommand{\eps}{\epsilon}

\newcommand{\EE}[1]{\underset{#1}{\mathbb{E}}}

\usepackage{bbm}

\providecommand{\bm}{\bm{m}}

\providecommand{\bu}{\bm{u}}
\providecommand{\bv}{\bm{v}}

\providecommand{\bx}{\bm{x}}
\providecommand{\by}{\bm{y}}
\providecommand{\bz}{\bm{z}}

\providecommand{\bmu}{{\bm{\mu}}}
\providecommand{\bsigma}{\bm{\sigma}}

\providecommand{\or}{\bar{r}}

\providecommand{\ox}{\bar{x}}

\providecommand{\oZ}{\bar{Z}}

\providecommand{\obz}{\bar{\bz}}

\renewcommand{\rw}{{\normalfont \textsf{w}}}
\renewcommand{\rx}{{\normalfont \textsf{x}}}
\renewcommand{\ry}{{\normalfont \textsf{y}}}

\providecommand{\defeq}{\coloneqq}

\newcommand{\cJ}{\mathcal{J}}

\newcommand{\cN}{\mathcal{N}}

\renewcommand{\eqref}[1]{equation~(\ref{#1})}

\renewcommand{\P}[1]{\mathbb{P} \left( #1 \right)}

\newcommand{\Exp}[1]{\exp \left\{ #1 \right\}}

\providecommand{\nats}{\mathbb{N}}
\providecommand{\reals}{\mathbb{R}}

 \newcommand{\lrc}[1]
    {\left\{ #1 \right\}}
    
    \newcommand{\lrp}[1]
    {\left( #1 \right)}

\providecommand{\lrbrace}[1]{\left\{ #1 \right\}}

\DeclareMathOperator{\rank}{rank}
\renewcommand{\inner}[2]{\left\langle #1, #2  \right\rangle}
\newcommand{\phase}[2]{\angle \left( #1, #2 \right)}
\providecommand{\abs}[1]{\left| #1 \right|}

\usepackage{times}

\newcommand{\Name}[1]{\textbf{#1}}  % Make the name bold
\newcommand{\Email}[1]{\texttt{#1}}  % Create a mailto link for emails

\title{\Huge Batch Normalization Decomposed\vspace{0.2cm}}

\author{
  \begin{tabular}{@{}l@{\hskip 3cm}r@{}}
    \Name{Ido Nachum\thanks{These authors contributed equally to this work.}} & \Email{inachum@univ.haifa.ac.il} \\
    \Name{Marco Bondaschi\footnotemark[1]} & \Email{marco.bondaschi@epfl.ch} \\
    \Name{Michael Gastpar} & \Email{michael.gastpar@epfl.ch} \\
    \Name{Anatoly Khina} & \Email{anatolyk@tauex.tau.ac.il} \\
  \end{tabular}
}
\date{}

\begin{document}

  \maketitle

  \vspace{1cm}
\begin{abstract}

% On the Geometrical Representation Induced by Batch Normalization.
\medskip

\emph{Batch normalization} is a successful building block of neural network architectures. Yet, it is not well understood. A neural network layer with batch normalization comprises three components that affect the representation induced by the network:  \emph{recentering} the mean of the representation to zero, \emph{rescaling} the variance of the representation to one, and finally applying a  \emph{non-linearity}. Our work follows the work of Hadi Daneshmand, Amir Joudaki, Francis Bach [NeurIPS~'21], which studied deep \emph{linear} neural networks with  only the rescaling stage between layers at initialization. In our work, we present an analysis of the other two key components of networks with batch normalization, namely, the recentering and the non-linearity. When these two components are present, we observe a curious behavior at initialization. Through the layers, the representation of the batch converges to a single cluster except for an odd data point that breaks far away from the cluster in an orthogonal direction. We shed light on this behavior from two perspectives: (1) we analyze the geometrical evolution of a simplified indicative model; (2) we prove a stability result for the aforementioned~configuration.

\end{abstract}
\newpage
\section*{Introduction}

% {\todo{Add a motivating figure with 4 subfigures: no BN, full BN, Bach-style BN, our BN.}}

% Maybe our model is not entirely far fetched???? Paper about convolutions and training 10000 layers...

A neural network (NN) consists of a sequence of \textit{transformations} or \textit{layers} sharing a similar structure.
% We consider each of these transformations as a layer of the network. 
Typically, the $t$-th transformation $t$-th transformation comprises two steps:

% \AK{Now this part is not notation-compatible with the sequel, say with Eq.~(1)... Also, in the sequel, we denote vectors by boldface letters but not in the intro for some reason. Is this to avoid confusion between column vectors ($x_1^{(t)}$ here) and the row vectors in Sec.~3 say?}
\begin{enumerate}
    \item Multiply the output of the previous layer $\bx^{(t-1)}$  by a matrix $W^{(t)}$:  ${\bz}^{(t)}=W^{(t)} \bx^{(t-1)}$.
    \item Apply a non-linearity $F:\RR\rightarrow\RR$ coordinatewise to attain the new output: $\bx^{(t)}=F\lrp{ {\bz}^{(t)}}  $.
\end{enumerate}
 
% A standard neural network is a sequential application of matrix multiplication that is followed by a nonlinearity. The output of a layer is determined in two steps:
In practice, many different tweaks are applied to these transformations to improve performance. Batch normalization (BN) \cite{bn} is a prime such example. It is a ubiquitous component of NNs as it decreases the training time and the generalization error~\cite{he,huang, silver} . A noteworthy aspect of BN is that the output corresponding to a single input depends on other inputs in the same batch.

% It  plays an important role in many neural network architectures~\cite{he,huang, silver}. 
Let  $X^{(t)}=\lrp{\left. \bx^{(t)}_1 \right\vert \cdots\middle| \bx^{(t)}_n}$ be the output of layer $t$ consisting of $n$ column vectors that correspond to a batch of size $n$. 
% Now, we  can 
Then, the output of every layer of a NN with BN is defined recursively as follows.
%  Let  $A^{(t)}=\lrp{a^{(t)}_1|\cdots|a^{(t)}_n}$ be the output of layer $t$ consisting of $n$ column vectors which correspond to $n$ inputs. An important 
% With BN, 
% with $T$  layers $\lrp{ W^{(1)},\cdots,W^{(T)}}$ and
% BN comprises two stages performed on each coordinate in the batch separately. 
% The output of a layer is determined as follows:
\begin{enumerate}
\item 
    Multiply the output of the previous layer $X^{(t-1)}$ by a matrix $W^{(t)}$: 
  \begin{align} 
{Z}^{(t)}=\lrp{  { \bz}^{(t)}_1 \middle\vert \cdots \middle| {\bz}^{(t)}_n} \coloneqq W^{(t)} X^{(t-1)} .
  \end{align}
  % \AK{This is confusing. Did you intend to denote the columns by $x_i^{(t)}$?}

    \item Compute the empirical mean and standard deviation
\begin{align}
 \label{eq:mean}
     \bmu^{(t)} &\coloneqq \frac{1}{n} \sum_{i=1}^n {\bz}^{(t)}_i ,
     &\bsigma^{(t)} &\coloneqq \sqrt{ \frac{1}{n} \sum_{i=1}^n \left({ \bz}^{(t)}_i- \bmu ^{(t)} \right)^2 } \:.
 \end{align}
    
    % \item Empirical standard deviation computation: 
% \begin{align}
  % \bsigma^{(t)}& \coloneqq \sqrt{ \frac{1}{n} \sum_{i=1}^n ({ \bz}^{(t)}_i- \bmu ^{(t)})^2 }
% \end{align}\label{eq:var}

%  X^{(t+1)}     &= F \lrp{\frac{ y^{(t)} -    \mu^{(t)}  }{ \sigma^{(t)} }     }  
% \end{align}
\item 
    Normalize $Z^{(t)}$ by \textit{recentering} (\textit{RC}) and \textit{rescaling} (\textit{RS}):
    \begin{align}
        \oZ^{(t)} 
        = \lrp{  \obz^{(t)}_1 \middle\vert \cdots \middle| \obz^{(t)}_n}
        \defeq \frac{ {Z}^{(t) }-    \bmu^{(t)}  }{ \bsigma^{(t)}     } 
        = \lrp{  \frac{\bz^{(t)}_1 - \bmu}{\bsigma} \middle\vert \ \cdots \ \middle| \frac{\bz^{(t)}_n - \bmu}{\bsigma}} .
    \end{align}
\item 
    Apply a non-linearity (NL) $F:\RR\rightarrow\RR$ coordinatewise to 
    the normalized batch $Z^{(t)}$:%\footnote{
    % $\bmu^{(t)}$ and $\bsigma\tu$ are column vectors so ${Z}^{(t) }-   \bmu^{(t)} $ means that we subtract $ 
 % \bmu^{(t)}$ from each column in  $Z^{(t) }$. Dividing by $\bsigma^{(t)} $ means that we divide each column of ${Z}^{(t) }-    \bmu^{(t)} $ coordinatewise by $\bsigma^{(t)} $.}
    % that includes a \emph{recentering} (\textbf{RC}) from eq.~\ref{eq:mean}  and 
% \emph{rescaling} (\textbf{RS}) from eq.~\ref{eq:var}: 
   \begin{align}
        X^{(t)} \coloneqq F \lrp{ \oZ^{(t)}}. %F\lrp{ \frac{ {Z}^{(t) }-    \bmu^{(t)}  }{ \bsigma^{(t)}     }  }  
   \end{align}

    \end{enumerate}

After applying BN during initialization (and training), the distribution of the preactivation values of each neuron has zero mean and unit variance (both during initialization and training). This was the original motivation for BN, namely, to reduce the rate at which the representations through the layers change, or put differently, to reduce the covariate shift~\cite{bn}. Yet, this is not the reason for its success~\cite{santukar}, and BN is still not well understood. 

In this paper, we follow \cite{bach20,bach21} where fully-connected \textit{linear} (no ReLU, sigmoid, etc.) neural networks were studied with only one part of BN applied---RS (part IV of Figure~\ref{fig:compare}).
Under that framework at initialization, it was shown that a full-rank batch becomes orthonormal asymptotically as the depth of the network grows.

%  Before presenting our contribution,   we mention some limitations of~\cite{bach20,bach21}: 

% \begin{itemize}

% \item  Analyzing a network without MS and NL.
    
%   \item See  Figure~\ref{fig:hist} for the difference between the output histograms of neurons with and without mean subtraction. 

%      \item A batch of full rank is assumed for the batch to reach orthonormality.
%     \end{itemize}

\vspace{-0.1cm}

\subsection*{Our Contribution}

\vspace{-0.1cm}

\begin{figure}[t!]
    \centering
    \vspace{-1.9cm}
    \includegraphics[trim=0 200 0  0, clip,scale=0.096]{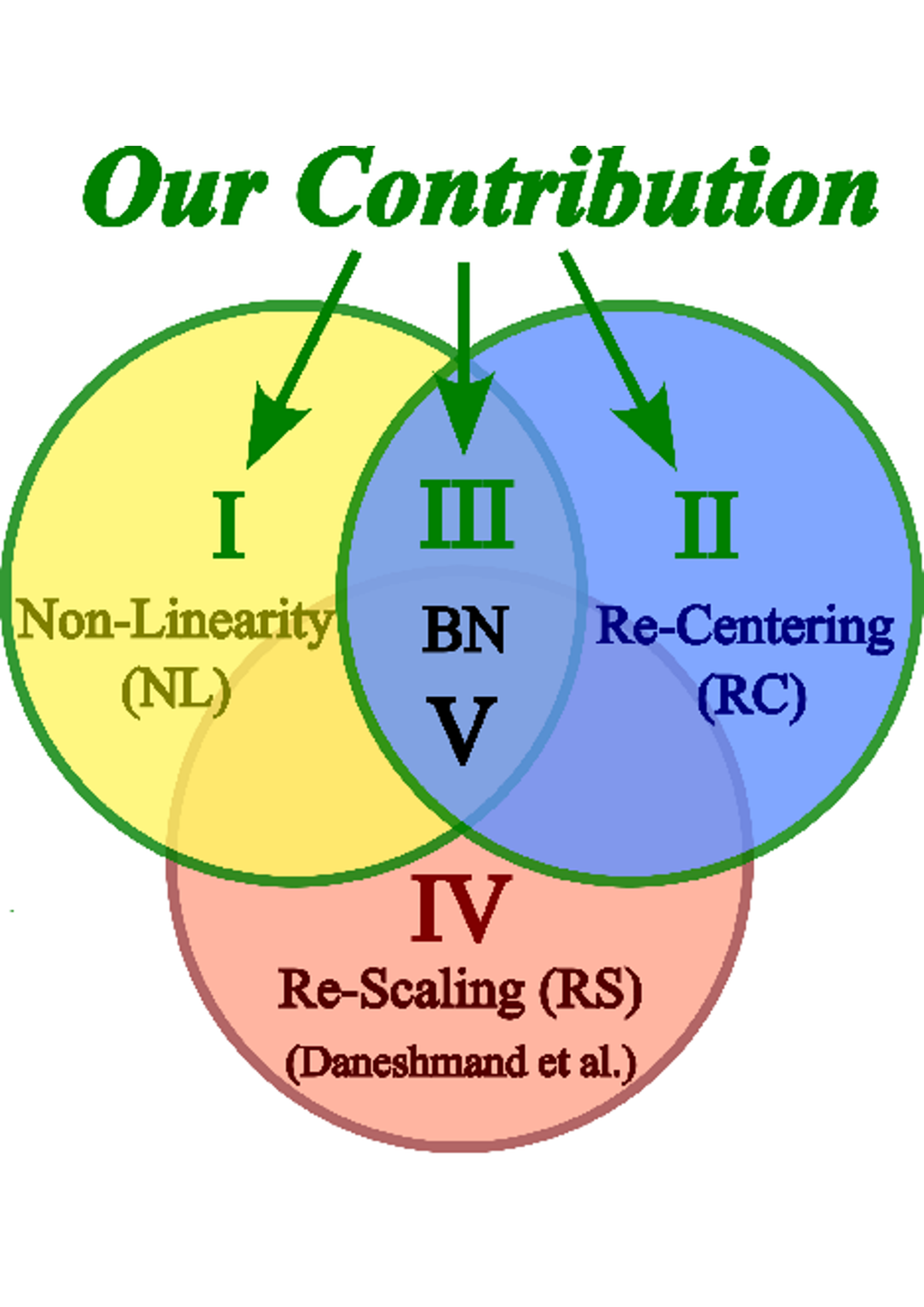} 
    \qquad
    \begin{tabular}[b]{|c | c |  c |}\hline
      BN Components &  Daneshmand et al.  & Our Contribution \\ \hline \hline 
    \textbf{I} \hfill   NL & \xmark & \cmark\\ \hline
    \textbf{II} \hfill   RC & \xmark & \cmark\\ \hline
 \textbf{III} \hfill   NL + RC & \xmark & \cmark\\ \hline
    \textbf{IV} \hfill   RS & \cmark &  \xmark\\ \hline
       \textbf{V } \hfill  ~~~ NL + RC +  RS  & \xmark & \xmark\\\hline
    \end{tabular}
    \captionlistentry[table]{A table beside a figure}
    % \captionsetup{labelformat=andtable}
    \caption{A comparison between previous work and our contribution. Our contribution studies the effects of the ReLU non-linearity and recentering at initialization and how they interact.}
    \label{fig:compare}
\end{figure}

\begin{figure}[t!]
    \centering
        \vspace{-0.2cm}
    \subfigure[Final training accuracy]{
        \centering
        \includegraphics[width=0.44\textwidth]{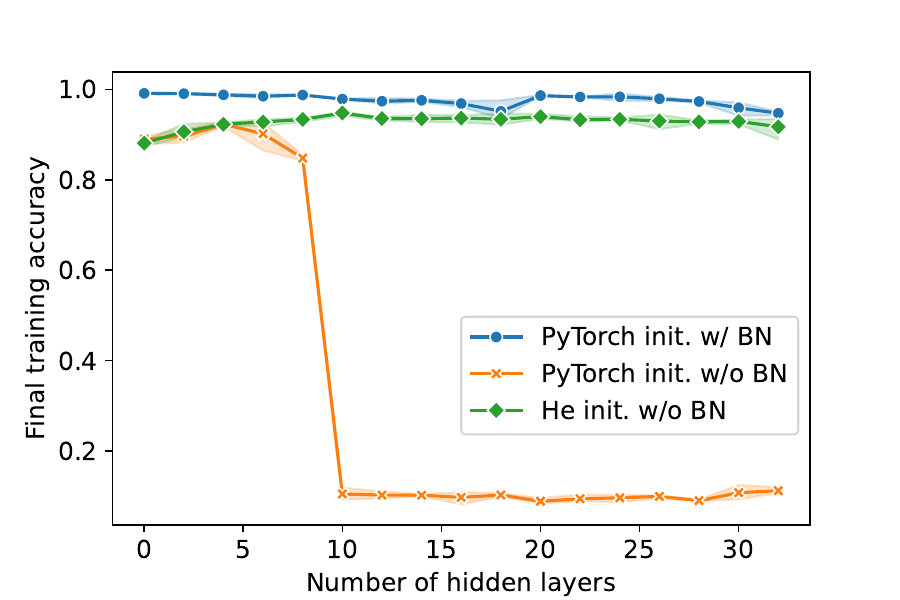}
        \label{fig:acc_init}}
    %\hfill
    \subfigure[Rank of the last hidden layer]{
        \centering
        \includegraphics[width=0.44\textwidth]{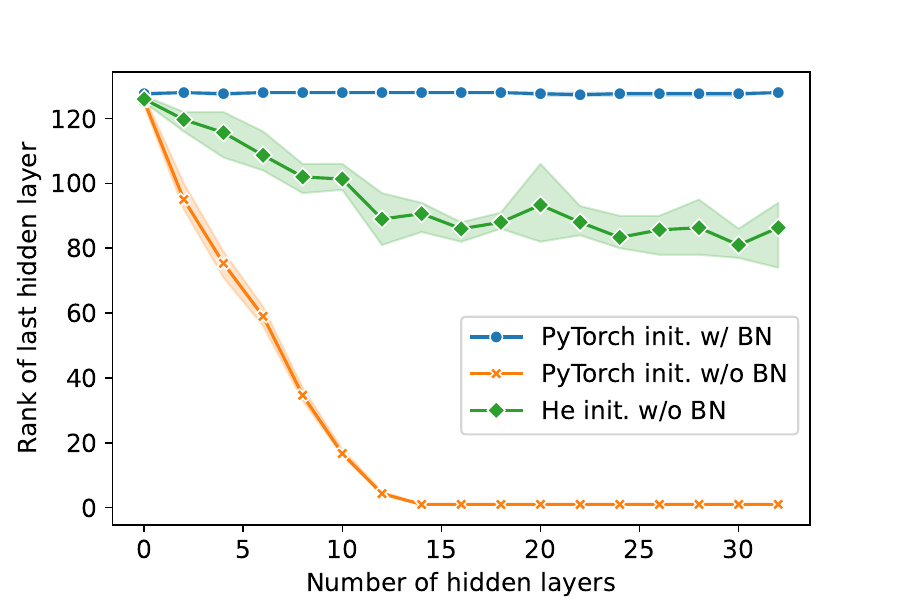}
        \label{fig:rank_init}}
    \caption{Comparison of final training accuracy and the rank of the last hidden layer in a fully-connected ReLU network 
    % for three cases
    using the supplementary code of~\cite{bach20}: (1)  with BN (2) without BN  (3)  without BN while changing \textbf{only} the default PyTorch initialization in the code to the He initialization.}
    \label{fig:init}
    \vspace*{-1em}
\end{figure}

Our contribution begins with a reference to Figure~1 in ~\cite{bach20}. That figure suggests that to achieve low training error, NNs must maintain a high rank when representing the training set through the different layers already at initialization (NNs may form low-rank representations at initialization, see~\cite{Saxe14exactsolutions}). Yet, this implication does not hold; see \figref{fig:init}:  a NN may reach high training accuracy from low-rank representations at initialization. To demonstrate this, we set the appropriate He initialization~\cite{kaiming} (PyTorch's default variance in the code of~\cite{bach20} is too small). \Figref{fig:init} clearly shows that the success of BN does not stem from a high-rank representation that it induces at initialization and that a more fine-grained approach is required to better understand BN. 

Indeed, as we demonstrate in \figref{fig:no_mean}, with NL the representation achieved by the hidden layers is not orthogonal.
% ; see Figure~\ref{fig:no_mean}. 
The more so, when applying BN without RC, the typical angle between a pair of vectors in say $\bx_1^{(30)}, \ldots ,\bx_n^{(30)}$ (the representation after 30 layers) is approximately $60^{\circ}$---as is evident from Figure~\ref{fig:angle_no_mean}---as opposed to a typical angle of $75^{\circ}$ in the case of BN---as is demonstrated in Figure~\ref{fig:angle_mean}. A possible explanation of this difference is offered by Figure~\ref{fig:hist}: when BN without RC is used, the histogram of the activity of some neurons in the network looks like the one shown in Figure~\ref{fig:hist_no}, whereas standard BN with RC induces the same type of histogram as in Figure~\ref{fig:hist_with} for almost all neurons; untypical histograms, like the one represented in Figure~\ref{fig:hist_no}, induce stronger correlations between inputs and hence the angle between vectors decreases.

 % together with NL and RS (complete BN), we notice that it has an effect on the batch representations through the layers:  vs. $75^{\circ}$ with RC (see Figure~\ref{fig:hist_mean}).

To understand the effect of each stage of BN, 
% That is why 
we decompose BN into its three different components (NL, RC, and RS; see Figure~\ref{fig:compare}), and complement the study of \cite{bach20,bach21} of RS only, by understanding the role of RC, NL, and how they interact. In correspondence with Figure~\ref{fig:compare}, we focus in Section I on the isolated effect of the ReLU NL on the rank of the representations, where we show that the rank of the representation increases substantially after one layer; in Section II, we focus on the isolated effect of RC and show that in linear networks it affects only the  representation of the first layer; in Section III we study the representation induced when combining the randomness of the initialization, NL, and RC; and in the discussion section, we contemplate how all the different pieces of the BN puzzle fit together and  suggest an alternative initialization scheme that was inspired by this theoretical work. 

% \AK{Maybe add as entence about the rank and stability results in the short descriptions of each section above?}

\begin{figure}[t!]
\centering
\subfigure[2D projection of the representations.]{
    \centering
    \includegraphics[trim=0 400 0  450, clip,width=0.45\textwidth]{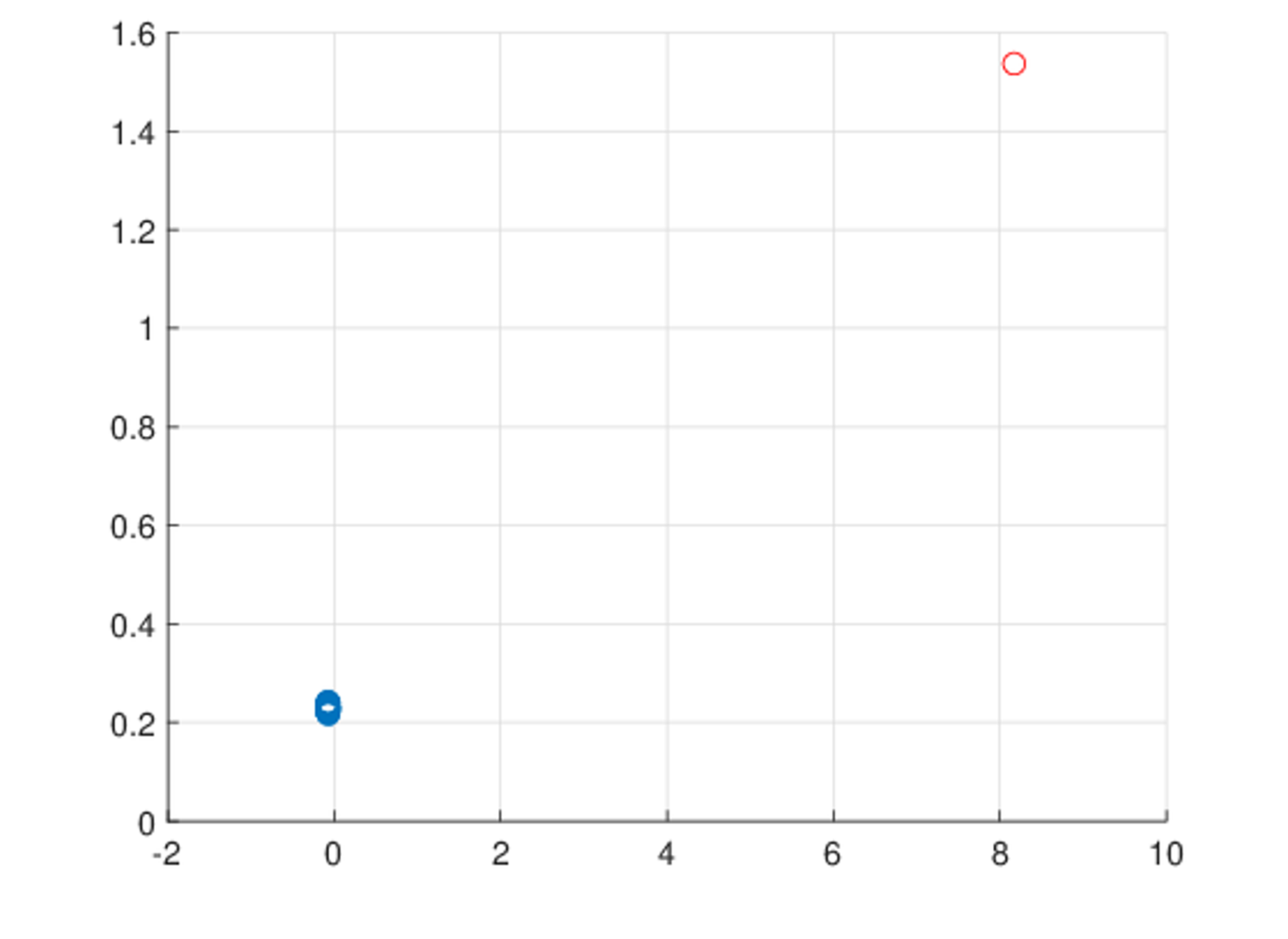}
    \label{fig:proj_ms}}
\subfigure[Relation between input and output angles.]{
    \centering
    \includegraphics[trim=0 500 0  450, clip,width=0.43\textwidth]{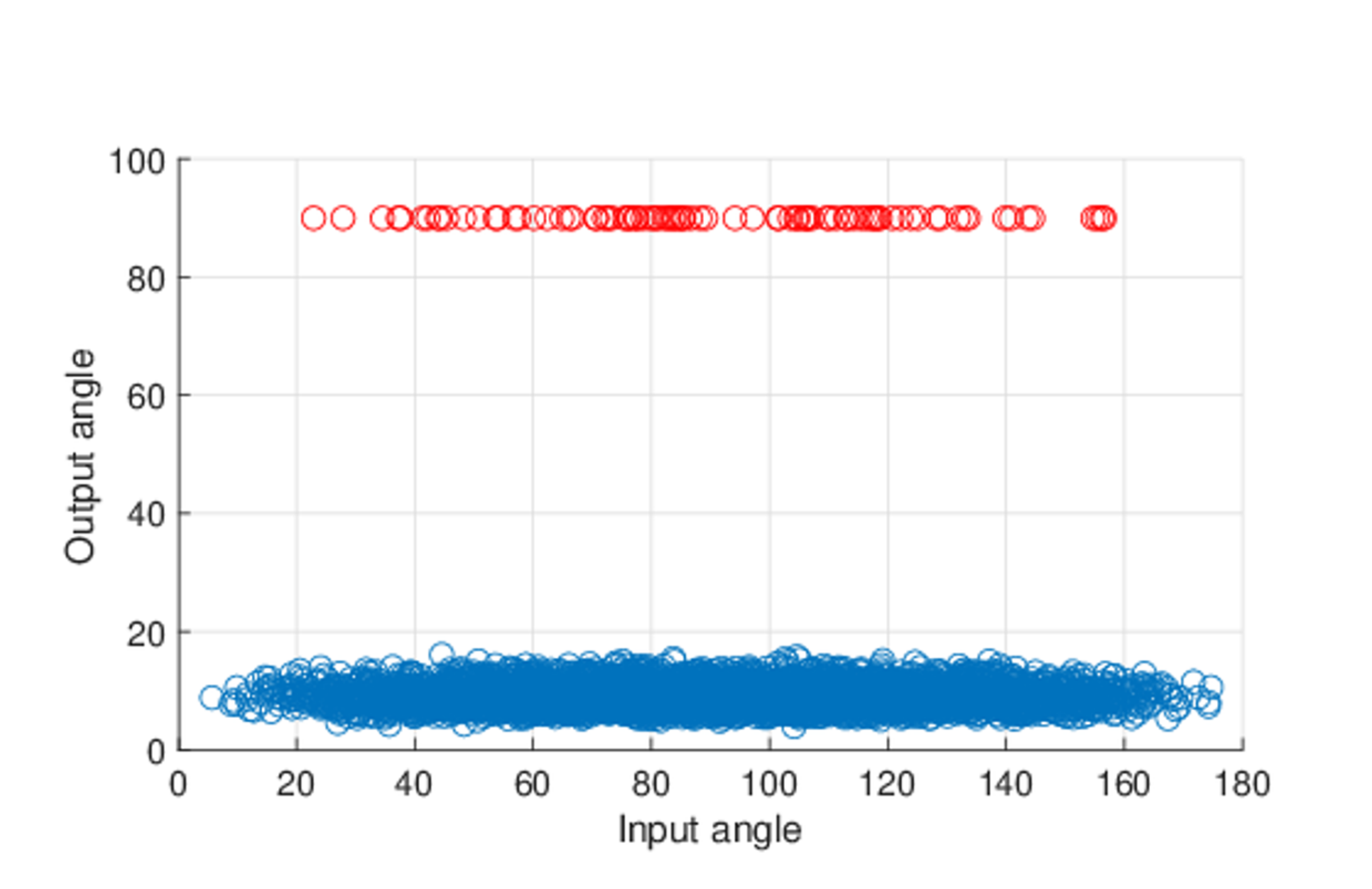}
    \label{fig:angle_ms}}
\centering
\caption{The batch representation induced by the final hidden layer with RC and ReLU NL. Figure (a) is a random two-dimensional projection of the final layer's representations. The "escaped" point is marked in red. Figure (b) represents the angles between pairs of vector representations before and after the final layer. The points marked in red represent the angles between the "escaped" point and any other point of the batch.}
\label{fig:geom_ms}
\vspace*{-1.2em}
\end{figure}

Section III is our main contribution, and in its setting, we observe a curious behavior: after enough layers, all the data points in the batch collapse to one point, except for a single odd data point that escapes far away from the cluster in an orthogonal direction. This behavior is illustrated in Figure~\ref{fig:geom_ms}. On a high-level, such behavior may explain the typical histograms that appear for all neurons, as in Figure~\ref{fig:hist_with},  on which we will elaborate further in the Discussion section. We explain the observation above via two theorems:
\begin{itemize}
    \item  Theorem~\ref{ThmAngle} explains the behavior above for a simplified model that is more tractable:
    % for better insight. 
    At each layer, instead of calculating the expected behavior over the typical Gaussian distribution over the weights, we analyze the expected  behavior over a simpler distribution. We show that this simplified model captures the  empirical behavior well.

    \item  Theorem~\ref{thm:rand_stb} deals with Gaussian random layers and provides a stability result. It states that an initial configuration that consists of a cluster and one separate data point, as above, remains geometrically unchanged after the application of a randomly initialized network with RC and ReLU activation.
    % behavior over the standard basis and its mirror axes $\{e_1,...,e_d,-e_1,...,-e_d\}$.
\end{itemize}
\textbf{All the proofs of the theorems can be found in the Appendix.} 
% On the one hand,   complete BN induces more symmetric representations (histograms of activations all look the same for all neurons) when compared to BN without RC. On the other hand, when applying only NL+RC (BN without RS) we get quite an unsymmetric representation.

% That is why we conclude our theoretical contribution with the complement case study to~\cite{bach20,bach21} (see Figure~\ref{fig:compare}), that is, matrix product followed by RC and ReLU NL. 

% Together with the ReLU non-linearity, this is not required.

% \AK{We need to talk first (perhaps in the main text of the intro.) about the clustering phenomenon for multi-dimensional inputs. Then say that we offer a tractable analysis for one-dimensional inputs and $W=I$, along with stability results for one-dimensional inputs with a random $W$.}

\begin{shaded*}
\centerline{\textbf{Notation}}
    \noindent Matrices are denoted by capital letters such as $X$. For a matrix $X$, its $i$-th row is denoted by $\rx_i$, and its $i$-th column by $\bx_i$. Element $(i,j)$ of $X$ is denoted by $X_{ij}$. The $i$-th element of a vector $\bm{y}$ is denoted by $y_i$. 
            
    \noindent The result of the operation denoted by $X - \bm{\mu}$, where $X$ is a $d\times n$ matrix, and $\bm{\mu}$ is a $d \times 1$ column vector, is understood to be equal to
    a $d \times n$ matrix $Z$ whose elements are $Z_{ij} = X_{ij} - \mu_i$. Similarly, the result of the operation $\frac{X}{\bm{\sigma}}$, where $X$ is a $d\times n$ matrix, and $\bm{\sigma}$ is a $d \times 1$ column vector, is understood to be equal to
    a $d \times n$ matrix $Z$ whose elements are $Z_{ij} = \frac{X_{ij}}{\sigma_i}$.
    \noindent The result of the operation $F(X)$, where $X$ is a matrix and $F: \mathbb{R} \to \mathbb{R}$ is a function, is a matrix $Z$ with elements $Z_{ij} = F(X_{ij})$.
   \end{shaded*}

\subsection*{Related Work}

 We mention some related works that attempt to understand the underlying principle of BN both theoretically and empirically. 
 
 \cite{karakida, arora2018, bjorck, lyu2022understanding} studied some beneficial properties of BN to explain its success,
\cite{frankle2020training} trained NNs where only the tunable parameters of BN are trained, achieving surprisingly good results,   and \cite{lubana2021beyond} tried to underpin some of the qualities of normalization layers in general. 
 
 \cite{yang2019mean} studied Deep NNs with BN and without residual connections and explained why they are hard to train; \cite{furusho2020theoretical}  and \cite{de2020batch} further elaborated on the synergy between skip connections and BN. 

Empirically, \cite{wang2022understanding} highlighted some of the shortcomings of BN in the Transformers architecture, and \cite{li2019understanding} pointed out some of the shortcomings of BN when used with dropout.

% \section*{BN Component I. ~ReLU Increases Rank}
\section*{I. ~ReLU NL Increases Rank}

\cite{bach20,bach21} study \textit{linear} networks (i.e., without NL) for which, to reach orthonormality of the vectors within the batch, the input batch needs to be assumed to be full rank. This is because the rank of the batch representations cannot increase through the layers, since $\rank(A  B) \leq \max\lrp{\rank(A), \rank(B) }$. 
In contrast, in a real setting, the NL increases the rank of the batch through the layers. We prove this phenomenon in Theorem~\ref{thm:rank}, which states that applying a random matrix followed by a ReLU NL over the training set induces full rank under a mild assumption. Hence, the NL provides a natural way to increase the dimension of the representations $X\tu$ of the batch at initialization.
% Theorem~\ref{thm:rank} states that a fully connected layer with the ReLU activation substantially increases the rank of any data set under the mild assumption that no vector in the batch is a multiple of the other.  
Theorem~\ref{thm:rank} uses the following geometrical quantity.

% We prove that the representation becomes full rank  with high probability, when using a layer with linear  amount of neurons  in the size of the batch $n$. The linear factor depends on the following geometrical quantity of the batch.

 \begin{definition}
 \label{eq:equivalence}
    Let $X=\lrp{\left. \bx_1 \right\vert \cdots \middle| \bx_n}\in\RR^{k\times n}$ be a  $k$-dimensional batch of size  $n$. We say that the row vectors $\rw_1,\rw_2\in \reals^k$ are equivalent, $\rw_1 \sim \rw_2$, if $\sign(\rw_1 \bx_i)=\sign(\rw_2 \bx_i)$ for all $1 \leq i\leq n$. We denote the finite set of  equivalence classes induced by this relation by $\{ \Gamma_l \}$ and define 
    \begin{align} 
        \gamma(X)= \min\{ \vol(\Gamma_l) :  \vol(\Gamma_l) > 0 \}
    \end{align} 
    where $\vol(\Gamma_l) = \P{\rw \in \Gamma_l}$ for $\rw \sim \mathcal{N}(0, I_k)$.
    % \EE{ w\sim \mathcal{N}(0, I_k)} 1_{\Gamma_l}\lrp{w}$.
 \end{definition}

\begin{exm}
Let $X=\lrp{\left. \bx_1 \right\vert \cdots\middle| \bx_n}$ such that $\bx_i= \lrp{\cos(2 \pi i/n), \sin(2 \pi i/n) }^T$ for $i \in [n]$ with odd $n$. That is, $X$ is composed of $n$ equally spaced points on the unit circle. It is clear that $\left| \{ \vol(\Gamma_l) :  \vol(\Gamma_l) > 0 \} \right|=n$  and  $\gamma(X)=1/n$.  %  $ $\vol(\Gamma_i)=\vol(\Gamma_j)$ for all $i,j$ with $\vol(\Gamma_j),\vol(\Gamma_j)>  0$

\end{exm}

To prove \thmref{thm:rank} below, we will use the following lemma.
\begin{lemma}
\label{lem:rank}
    Let $\bx_1, \bx_2, \ldots \bx_{t+1} \in \reals^k$ be column vectors such that no two vectors are collinear. 
    Denote $X^{(\ell)} \defeq \lrp{\bx_1 \middle| \bx_2 \middle| \cdots \middle| \bx_\ell}$ for $\ell \in [t+1]$.
    Denote further $W^{(r)} \in \reals^{r \times k}$, 
    and assume that $\rank \lrp{ \R \lrp{ W^{(r)} X^{(t)} } } = t$. 
    % Let further $\rw_{t+1} \sim \cN \lrp{0, I_k}$, 
    Then, with probability at least $\gamma \lrp{ X^{(t+1)} }$, 
    \begin{align}
    \label{eq:lem:rank}
        \rank \lrp{ \R \lrp{ W^{(r+1)} X^{(t+1)} } } &= t+1, %& \forall t \in [n] ,
    \end{align}
    where $W^{(r+1)} = \begin{bmatrix} W^{(r)} \\ \rw_{r+1} \end{bmatrix}$
    for $\rw_{r+1} \sim \cN \lrp{0, I_k}$.
    % For the setting of \thmref{thm:rank}, 
    % denote the matrix composed of the first $\ell$ \emph{columns} of $X$ by $X^{(\ell)}$.
    % Denote the $j$-th \emph{column} of $X$ by $X_j$, and the $i$-th \emph{row} of $W$ by $W_i$.
    % and the matrix composed of the first $\ell$ \emph{rows} of $W$ by $W^{(\ell)}$.
    % Then, for all $t+1 \in [n]$, 
    % Assume that $\rank \lrp{ \R \lrp{ W^{(r)} X^{(t)} } } = t$.
    % Then, %with probability of at least $\gamma \lrp{ X^{(t+1)} }$, 
    % \begin{align}
    % \begin{align} 
    % \label{eq:lem:rank}
    %     &\CP{ \rank \lrp{ \R \lrp{ W^{(r+1)} X^{(t+1)} } } = t+1}{\rank \lrp{ \R \lrp{ W^{(r)} X^{(t)} } } = t, W_i   X_j \neq 0 \ \forall i,j} 
    % \nonumber
    %  \\ &
    %  \qquad\qquad\qquad\qquad 
    %  \qquad\qquad\qquad\qquad 
    %  \qquad\qquad\qquad\qquad 
    %  \qquad\qquad
    %  \geq \gamma \lrp{ X^{(t+1)} }.
    % % \end{align}
    % \end{align}
\end{lemma}

\begin{theorem}
\label{thm:rank}
    Let $X\in\RR^{k\times n}$ be a  $k$-dimensional batch of size  $n$ and assume that no two columns of $X$ are collinear. %(very mild assumption)
    Let $W \in \reals^{\infty\times k}$ be  a  random matrix  with a countably infinite number of rows, each of size $k$ and  i.i.d.\ entries $\mathcal{N}(0,1)$. Define $W^{(d)}$ as the first $d$ rows of $W$, $y \defeq \min \lrbrace{d:\ \rank \lrp{ \R \lrp{W^{(d)} X}} =n}$, and $\gamma \defeq \gamma(X)$. %$r_d= rank(\R(W^{(d)} X ))$, and $Y=\min\{d: r_d=n \}$. Then, 
    Then, 
    \begin{enumerate}
    \item  
        $ \E{y} \leq n/\gamma$. 
      %  \AK{You defined $d$, so it's not clear what you are minimizing over...}
    
    \item 
        For $d = \alpha n / \gamma$ for some $\alpha > 2$, % and $\gamma \defeq \gamma(X)$, 
        \begin{subequations}
        \begin{align} 
        \!\!\!\!
        \!\!\!\!
            \P{\rank(\R(W^{(d)} X))=n} 
            &\geq  1 - \Exp{-n \lrp{ \frac{\alpha}{\gamma} \log \frac{1}{\alpha} + \lrp{\frac{\alpha}{\gamma} - 1} \log \frac{\alpha - \gamma}{1 - \gamma} }}
         \\ &\geq 1 - \Exp{-n \lrp{ \alpha - 1 - \log \alpha }}.
        \end{align} 
        \end{subequations}
        % $\rank(\R(W X))=n$
        % with a probability of at least $ 1- 2^{- n (2+\alpha) \lrp{1-h\lrp{\frac{1}{2+\alpha  }} }+1}$, 
      
        % where $h$ denotes the binary entropy function, which is defined as $h(p) \defeq -p \log_2(p) - (1-p) \log_2(1-p)$.
    \end{enumerate}
\end{theorem}

% \begin{rem}
% The inequality in item 1 is tight, up to a factor of $3$, for  $X=\lrp{\left. \bx_1 \right\vert  \bx_2 \middle| \bx_3}$ where $\bx_1=(1,0)$,  $\bx_2=(\cos(\gamma),\sin(\gamma))$, and $\bx_3=(\cos(2\gamma),\sin(2\gamma))$ for $0<\gamma\ll 1$. \AK{Why is this case interesting? Because of the example? If so, this should be clearly stated...}

% \red{LEAVE OR REMOVE?--->} More so, one may derive a tighter bound that is tight for many other scenarios, using a more sophisticated  quantity than $\gamma$. In the previous example, where all data points are distributed uniformly on the sphere, we can arrive at $\E{ Y }\leq  n \log (n)$.

% This will require a treatment of a generalized coupon collector's problem: we have $T$ coupons $\{c_i\}_{i=1}^T$,   each occurring with probability $p_i$.  What is the expected number of draws to get $n\leq T$ coupons? 
% \end{rem}

The theorem states that at most $n/\gamma(X)$ ReLU neurons are required, in expectation, to guarantee that a representation of any batch will be full rank (item 1), and the probability of not achieving this decays exponentially fast in $\alpha$ for $\frac{\alpha n}{\gamma}$ neurons (item 2). The theorem still holds when BN is applied before ReLU, as stated in the next corollary. The proof of the corollary is left to the reader.
% Corollary~\ref{cor:rank}.

\begin{corollary}
\label{cor:rank}
          Theorem~\ref{thm:rank}  holds for    $y \defeq \min \lrbrace{d:\ \rank \lrp{ \R \lrp{BN\lrp{ W^{(d)} X}}} =n}$, where $BN$ refers to recentering and rescaling and $\gamma\coloneqq \gamma\lrp{X-\frac{1}{n} \sum_{i=1}^n \bm x_i}$.
\end{corollary} 

Let us now compare Theorem~\ref{thm:rank} to Lemma~5 in \cite{sv_NN}. The authors generalize the notion of singular values of linear operators to ReLU neural layers and show that the singular values do not increase for such ReLU operators. 
Theorem~\ref{thm:rank} shows that such a generalization does not capture the spirit of singular values for the following reason. The number of non-zero singular values of an operator equals the dimension of its image. So, for example, by Lemma~5 in~\cite{sv_NN}, for a linear operator followed by ReLU with input dimension $2$, the maximal number of non-zero singular values is $2$. This would suggest that the image dimension of the operator is at most $2$. Theorem~\ref{thm:rank} shows that, in fact, the image dimension scales linearly with the number of rows, so there should be many more non-zero singular values for a generalization of singular values of a ReLU layer.

 \section*{II.  ~Recentering}
 
For completeness, we refer briefly to the simple scenario where we use only RC over a linear network. When using RC alone without NL, the representation changes only at the first layer, where the mean is set to zero in every coordinate, and this is maintained through the layers because the mean of every coordinate remains zero  after applying the following linear layers. That is, if $\sum_i \bx_i = \bm{0}$,  then $\sum_i W \bx_i =\bm{0}$ for any $W$. Therefore, when used alone, RC does not have any effect on the batch representation after the first layer.
 
 \section*{III.  ~Recentering Followed by ReLU NL}

In contrast to the previous section, RC has an effect through all the layers when it is followed  by a ReLU NL. Figure~\ref{fig:geom_ms} displays a behavior that is consistent for all the batches that we experimented with: after enough layers, all of the data points collapse to a single cluster, and one \emph{odd} data point  deviates far away,  in an orthogonal direction to the cluster.

Let us give a high-level intuition about why the representation in Figure~\ref{fig:geom_ms} emerges.  Assume there is one data point, say $\bx_1$, that is far, to some degree, from the mean of the rest of the datapoints, namely, from the center of the cluster $\bm \nu \coloneqq \frac{1}{n-1}  \sum_{i=2}^n \bx_i $. Over each layer, we expect the following to occur:
\begin{enumerate}
   
    \item \label{it1} We project the datapoints by  a random $\rw$ and expect the distance between the projected $\bx_1$ and the projected $\bm \nu $  to remain large.

    \item  \label{it2} We then subtract the projected mean of all datapoints $ \mu \coloneqq \frac{1}{n}  \sum_{i=1}^n \rw \bx_i $. \\ 
    If $\rw \bm \nu  - \mu <0$, then $\rw \bx_1  -  \mu >0$ and it is 
    likely that many points in the cluster now have negative entries.  \\
    If $\rw \bm \nu  - \mu >0$, then $\rw \bx_1  -  \mu <0$ and it is likely that only few points in the cluster now have  negative entries.

    \item  \label{it3} We apply ReLU to the datapoints. \\
    If $\rw \bm \nu  -   \mu <0$, then the points in the cluster  with negative entries all map to zero. We get a tighter cluster, while the value of the projected odd point does not change. \\
If $\rw \bm \nu  -  \mu >0$,  then $\bx_1$ maps to zero with a few points from the cluster and the values of other points do not change.

\end{enumerate}

In total, the effect on the representation of a projection with $\rw \bm \nu  -  \mu >0$ is much greater than the case with $\rw \bm \nu  -  \mu <0$ since in the former case much more points map to zero compared to a selected few in the latter (so the geometry does not change much on average). 
Hence, when moving from layer to layer, 
% if we keep scaling the norm of the representation of $\bx_1$ to $1$, 
we expect the cluster to become tighter and move closer to zero fast, and the ratio between the representation of $\bx_1$ and the mean of the cluster to become large. This effectively  reproduces the geometry in Figure~\ref{fig:geom_ms}.

A typical approach for explaining such behavior would follow two steps. First, for a single ReLU neuron with a corresponding row of weights $\rw$, calculate what would happen in expectation for the representation of the batch. Secondly, show that this behavior is concentrated around that mean when using many neurons.

For calculating the expectation, we are required to compute
\begin{align}\label{eq:dual_act}
    \EE{\rw\sim \mathcal{N}(0,I_d)}  \R\lrp{\rw \lrp{ \bx_i-\frac{1}{n}\sum_{r=1}^n{\bx_r } } } \R \lrp{ \rw \lrp{ \bx_j -\frac{1}{n}\sum_{r=1}^n\bx_r}  } .
\end{align}
This is the expected inner product between inputs $\bx_i$ and $\bx_j$ after a single layer, which appears to be intractable while keeping tabs over the mean  $\frac{1}{n}\sum_{r=1}^n{\bx\tu_r }$ from layer to layer. This is why we separate our analysis into two theorems. Theorem~\ref{ThmAngle} shows how the representation in Figure~\ref{fig:geom_ms} emerges under a simplified model, and Theorem~\ref{thm:rand_stb} shows that such a representation remains invariant under the typical setting, as in equation~\ref{eq:dual_act}.

 % For a symmetric data set, we encounter a problem; since the operation of RC followed by ReLU is itself symmetric, there would be no symmetry breaking when calculating the expectation, and no data point would become odd.

\subsection*{Simplified Model for Recentering + ReLU}

We suggest a simplified model to explain the behavior of Figure~\ref{fig:geom_ms}. We start by noting that ReLU is homogeneous. That is, $\R(\alpha x)=\alpha \R(x)$ for $\alpha \geq 0$. Hence, instead of calculating the mean over all projections $\rw$ in $\RR^d$, as in \eqref{eq:dual_act}, we can calculate the mean over the unit sphere $\mathbb{S}^{d-1}$. 
Another simplification that we use is averaging over all standard unit vectors and their inverses $S_d\coloneqq \{e_1,\ldots,e_d,-e_1,\ldots,-e_d\}$ instead of averaging over all possible angles:

$$\EE{ \rw \sim \mathcal{N}(0,I_d)  }[\cdot ]  ~ \Longrightarrow ~ \EE{ \rw \sim U(\mathbb{S}^{d-1})  }   [\cdot ]   ~ \Longrightarrow ~   \EE{ \rw \sim  U \left(   S_d \right)  }  [\cdot ]  $$

At  first glance, the computation $ \EE{\rw  \sim  U \left(   S_d \right)  }  [\cdot ] $ may seem too simplistic. For example, if we start with a three-dimensional data set, we need to  sum  only 6 terms. Yet, when we follow this model through many layers, we get an exponential build-up in complexity. In fact, the output of every new layer of the network consists of all possible projections in $S_d$ of the output of the previous layer. Therefore, the number of summands doubles with each layer: if the batch representation has dimension $d$, then after one layer, its dimension increases to $2d$ (since $\lvert S_d \rvert = 2d$) after two layers---to $2^2 d$, and after $t$ layers---to $2^t d$.
% After one additional layer, it becomes $2 d$. 
% The process continues in the same way so that after $t$ layers, the dimension of the representation will equal  $2^t d$. 
In each step of the process, which corresponds to a layer, we project each dimension $i$ of the input representation onto the unit vectors $e_i$ and $-e_i$. This is why we can visualize this branching process as  a (perfect) binary tree, where each vertex of depth $t$ represents a neuron of the $t$-th layer of the network, see Figure~\ref{fig:tree} below. 

% project dimension $i$ of the input representation to two dimensions in the output representation -- can be visualized as a (perfect) binary tree, where each vertex of depth $t$ represents a neuron of the $t$-th layer of the network (see Figure~\ref{fig:tree} below). 

This model has a nice property: in contrast to the general model, which is induced by \eqref{eq:dual_act}, in the simplified case, we can easily keep track of the effect of each neuron in the network (or vertex in the tree) on the representation induced by each layer. This holds because the output of each neuron depends only on its parent neuron from the previous layer. This means that for any  branch
that begins on some vertex in the tree, its analysis is independent of all the other disjoint branches; this makes the analysis tractable.

That said, if we start with input dimension $d>1$, we will have $d$ disjoint trees with no interaction between them. So separating the dimensions would yield a different odd point on every coordinate, and no unique odd point can be identified from the batch. It is worth mentioning that no odd point can be identified for Gaussian matrices if our batch is symmetric; in this case, we must wait until the randomness, from layer to layer, will artificially produce an outlier. 

Consequently, we work in our model with input dimension $d=1$. That is, $X^{(0)}$ is a row vector with $n$ elements. In this case, we can identify the odd point. If $X^{(0)}$ is ordered, it would be either $x_1$ or $x_n$, and we can rigorously  prove  the high-level intuition, as in items~\ref{it1},~\ref{it2}, and~\ref{it3} above.

\begin{rem}
    Empirically, we observe that the odd point would typically be the one with the largest norm. This is compatible with the description above. So analogously, our Theorem~\ref{ThmAngle}  holds for  input dimension $d>1$ under  the condition that there are two points such that for each coordinate: (1) the entries of one point are greater than all other entries; (2) the entries of the second point are smaller than all other entries. 
\end{rem}

% In between the definitions and theorem statements, we will illustrate their role and interpretation in our model. 

% we have that $X^{(0)}$ is a row vector with $n$ elements

Let us now  introduce our model and let $X^{(t)}$ be the representation of the batch at layer $t$. Since the starting batch is made up of $n$ one-dimensional points, $X^{(0)}$ is a row vector with $n$ elements, and, due to the branching process described above, in general, $X^{(t)}$ is a matrix of size $2^t \times n$. Each of its columns, which we denote by $\bx_{i}^{(t)}$ for $i\in\{1,2,\dots,n\}$, is the representation of one data point at layer $t$ of the network. Each of its rows, which we denote by $\rx_j^{(t)}$ for $j\in\{1,2,\dots,2^t\}$, represents the output of one neuron at layer $t$ of the network. The projection of $X^{(t)}$ over all standard unit vectors and their inverses, i.e., over all vectors in $S_{2^{t+1}}$, is exactly equivalent to performing the matrix multiplication $W^{(t+1)} X^{(t)}$, where
\begin{align}
    W^{(t+1)} = \begin{pmatrix}
    +I_{2^t} \\
    -I_{2^t}
    \end{pmatrix}
    ,
\end{align}
and $I_{d}$ is the identity matrix of size $d$. In fact, for every $i\in\{1,2,\dots,2^t\}$, the output of neuron $i$, that is, the $i$-th row of $X^{(t)}$, generates two neurons' outputs at the next layer, one corresponding to the projection according to $e_i$, and the other according to $-e_i$. After the projections, RC, and the ReLU non-linearity are applied to each neuron's output. The result is the batch representation at layer $t+1$, $X^{(t+1)}$, which is therefore given by the equation
\begin{align}
\label{XtDef}
    X^{(t+1)} = \R\left(W^{(t+1)} \left(X^{(t)} - \bar{\bx}^{(t)}\right)\right) ,
\end{align}
where $\bar{\bx}^{(t)} = \frac{1}{n} \sum_{i=1}^n \bm{x}_i^{(t)}$.

It is clear from the given description that the evolution of each neuron's output from layer $t$ to $t+1$ does not depend in any way on the other neurons of that layer. Hence, we can focus on each neuron independently. In particular, we can say that, for every $i$, the output of neuron $i$, i.e., $\rx_i^{(t)}$, undergoes two different transformations, a \emph{positive} one, which consists of the projection according to $e_i$ followed by RC and ReLU, and a \emph{negative} one, which comprises the projection according to $-e_i$, RC and ReLU. The following is a formal definition of these two transformations.

\begin{definition}[positive and negative transformations]
We say that $\ry\in\mathbb{R}^n$ is the result of a \emph{positive transformation} applied on $\rx\in\mathbb{R}^n$ if $y_i = \R(x_i - \ox)$ for every $i\in [n]$, where $\ox = \frac{1}{n}\sum_{i=1}^n x_i$.
We say that $\ry\in\mathbb{R}^n$ is the result of a \emph{negative transformation} applied on $\rx\in\mathbb{R}^n$ if $y_i = \R(-x_i + \ox)$ for every $i \in [n]$.
\end{definition}

A visual representation of the described process as a binary tree is shown in Figure~\ref{fig:tree}, where each node of the tree represents the output of the corresponding neuron, and each edge connects each output with the two outputs generated at the next layer after the positive and negative transformations.

\begin{figure}[t]
\centering
\includegraphics[width=0.85\linewidth]{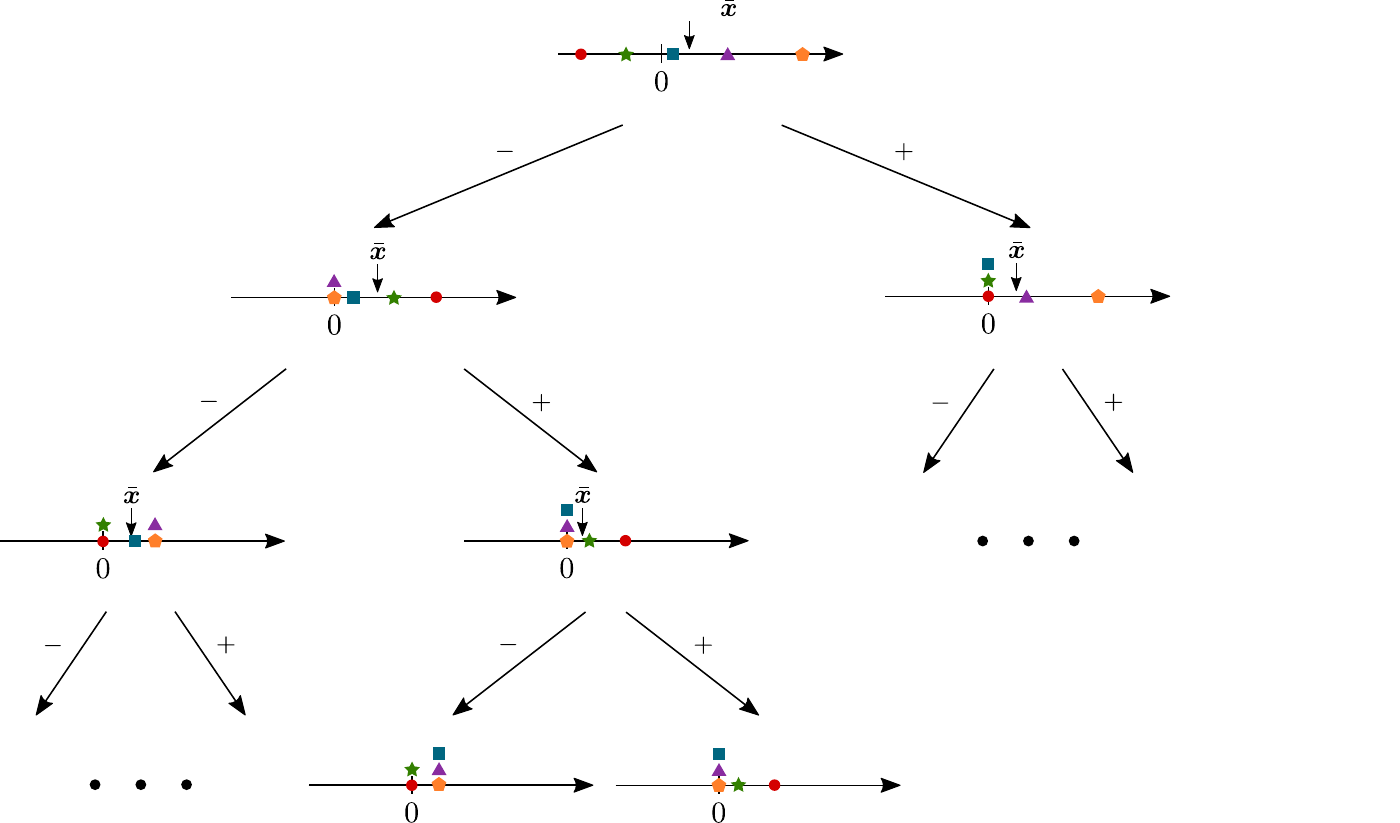} 
\caption{Partial example of the first three layers of the tree generated by the process of positive/negative transformations analyzed in this paper, starting from a one-dimensional batch with $n=5$ elements. Different elements have different shapes, to make it easier to follow their change of position. The average of the vector at each step is denoted by $\ox$.}
\label{fig:tree}
\end{figure}

The first result that we will present about our model is that the outputs of the neurons get more and more clustered as we go through the layers of the network. Therefore, the following definition will be useful.
% since the output of any neuron only depends 

% This makes our model tractable when following the contribution of each neuron upstream along the network.

\begin{definition}
    A vector $ \rx = (x_1,x_2,\dots,x_n) \in\mathbb{R}^n$ with possibly repeated entries is \emph{composed of $k$ clusters} if 
    % \IN{Toli: shouldn't C be just a set?}
    % \begin{enumerate}
    % \item   
    %     There exists an ordered set of $k$ real numbers $C = (c_1,\dots,c_k)$, with \\${c_1 < c_2 < \cdots < c_k}$, such that, for every $i\in\{1,2,\dots,n\}$, $x_i \in C$;
    % \item 
    %     There is no set $\tilde{C}$ with $\tilde{k} < k$ elements that satisfy the previous requirement.
    % \end{enumerate}
    % For a given vector $\bx$, we denote by $C(\bx) = (c_1(\bx),\dots,c_k(\bx))$ the unique set defined above, and we define $c(\bx) = \lvert C(\bx)\rvert$.
    it contains exactly $k$ unique entries. Denote the 
    % ordered 
    sequence of those unique $k$ entries of $\rx$ in ascending order by $C(\rx) \defeq (c_1(\rx), c_2(\rx), \ldots, c_k(\rx))$, and its length---by $c(\rx) \defeq \left| C(\rx) \right|$;
    clearly, $k \leq n$.
\end{definition}

The following result is immediate.
\begin{lemma}
\label{lem:c(f(x))<=c(x)}
    % Note that if any deterministic transform $f$ is applied to each element of the vector vector $\bx$ to obtain the vector $\by,$ then we must have $c(\by) \le c(\bx)$, since deterministic transforms cannot break up clusters.
    Let $y_i = f(x_i)$ for all $i \in [n]$ for some deterministic function $f$.
    Then, $c(\ry) \leq c(\rx)$.
\end{lemma}

Further, we will show that, as we proceed along the network layers, the number of clusters in each neuron's output will converge to either 2 or 3. Hence, the following definitions are natural.
\begin{definition}[stable]
A vector $\rx \in\mathbb{R}^n$ is \emph{stable} if $c(\rx)\leq 3$.
\end{definition}

\begin{definition}[more clustered]
A vector $\rx\in\mathbb{R}^n$ is \emph{more clustered} than a vector $\ry\in\mathbb{R}^n$ if $c(\rx) < c(\ry)$. %Conversely, we say that $\by$ is \emph{less stable} than $\bx$.
\end{definition}

With these definitions available, we are ready to present the first result of this section, on the clustering of the neurons' outputs.

%\todo{Add an explanation about the tree of positive and negative transforms and push the tree figure here. And explain that we are using the definitions for approximating random matrix with mean subtraction.}

\begin{theorem}
\label{ThmStable}
Suppose that a vector $\rx^{(0)}\in\mathbb{R}^n$ with all distinct components undergoes an infinite sequence of positive or negative transformations, resulting in the sequence of vectors $\{\rx^{(t)}\}_{t\geq 0}$, viz., for every $t \geq 0$, $\rx^{(t+1)}$ is the result of a positive or a negative transformation applied on $\rx^{(t)}$. Then, there exists a finite $t_0 \geq 0$ such that $\rx^{(t_0)}$ is stable. %composed of at most $3$ clusters (i.e., it is stable).
\end{theorem}

Essentially, Theorem \ref{ThmStable} says that, under our simplified model, there is a layer $t_0$ in the network after which each neuron's output, i.e., each row $\rx_i^{(t)}$ of the matrix $X^{(t)}$, for every $t\geq t_0$, is a vector composed of at most 3 clusters. This is because, at any layer, each neuron's output is generated by a sequence of positive and negative transformations starting from $X^{(0)}$.

Our next step is to study the asymptotic geometry of the batch as the number of layers increases. In particular, we want to study the behavior of the (normalized) inner product between any two datapoint representations, i.e., any two columns $\bx_i^{(t)}$ of $X^{(t)}$, when the number of layers goes to infinity. Formally, for any $i,j \in \{1,2,\dots,n\}$, we are interested in the quantity
\begin{align}
\label{AngleDef}
    \phase{\bx_i^{(t)}}{\bx_j^{(t)}} \coloneqq \arccos\frac{\inner{\bx_i^{(t)}}{\bx_j^{(t)}}}{\norm{\bx_i^{(t)}} \cdot \norm{\bx_j^{(t)}}}
\end{align}
in the limit of $t \to \infty$, where $\bx_i^{(t)}$ is the $i$-th column of $X^{(t)}$,
and $\inner{\cdot}{\cdot}$ and $\norm{\cdot}$ denote the standard Euclidean inner product and norm, respectively.
% for any $i\in\{1,2,\dots,n\}$. 
To that end, we prove the following theorem.

%\IN{Convinced now the index t should be in superscript $x^{(t)}$ }
\begin{theorem}
\label{ThmAngle}
Let $X^{(0)}\in\mathbb{R}^n$ be a row vector with all distinct components, and assume, without loss of generality, that its entries are increasingly ordered. Let $X^{(t)}$ be the $2^t \times n$ matrix defined in \eqref{XtDef} for $t \in \nats$, and denote its $i$-th column by $\bx_i^{(t)}$. 
Then, the following facts are always~true.
\begin{enumerate}
\item 
\label{itm:>=2_clusters}
    $\phase{\bx_1^{(t)}}{\bx_n^{(t)}} = \pi / 2$ for all $t \in \nats$.
\item 
\label{itm:<=3-clusters}
    $\lim\limits_{t \to \infty} \phase{\bx_i^{(t)}}{\bx_j^{(t)}} = 0$ for all $i,j\in\{2,\dots,n-1\}$.
\item 
\label{itm:norm_ratio}
    For any $i\in\{2,\dots,n-1\}$, either $\lim_{t\to\infty} \frac{\norm{\bx_i^{(t)}}}{\norm{\bx_1^{(t)}}} \leq \frac{3}{n-2}$ or $\lim_{t\to\infty} \frac{\norm{\bx_i^{(t)}}}{\norm{\bx_n^{(t)}}} \leq \frac{3}{n-2}$.
    % or
    % \begin{align}
        
    % \end{align}
\item 
\label{itm:almost_orthogonal}
    For any $i\in\{2,\dots,n-1\}$, either
    \begin{align}
        \frac{\pi}{2} - \frac{\sqrt{2}\pi}{n-2} 
        \leq \lim_{t\to\infty} \phase{\bx_i^{(t)}}{\bx_1^{(t)}}
        \leq \frac{\pi}{2} ~~~~ \text{or} ~~~~     \frac{\pi}{2} - \frac{\sqrt{2}\pi}{n-2} 
        \leq \lim_{t\to\infty} \phase{\bx_i^{(t)}}{\bx_n^{(t)}}
        \leq \frac{\pi}{2}.
    \end{align}
    % or
    % \begin{align}
    
    % \end{align}
    % \item exponential rate of convergence. \AK{Need to detail more...}
\end{enumerate}
\end{theorem}

\textit{Interpretation of the theorem.}
The theorem shows that, if a one-dimensional batch $X^{(0)}$ evolves according to our model, after a number of layers large enough, the vector representations of the data points converge to a configuration with the following properties: the representations of the largest and smallest starting datapoints become orthogonal (item 1 of the Theorem); the datapoint representations are all clustered together except for one or two points (item 2); one of the points "escapes" far away from the other clustered points, in such a way that the ratio between the norm of the escaped point and that of any point in the cluster is $\Theta(n)$ (item 3), and the angle between the escaped point and the cluster is approximately $90^\circ$ (item 4).

It is important to note that this behavior, derived from the simplified model described at the beginning of the section, is remarkably similar to what happens in the general case of a network with random weights, recentering, and ReLU non-linearity. This behavior, which we described in the introduction, is depicted in Figure~\ref{fig:geom_ms} above. Hence, our model, even if simpler (and easier to analyze), approximates well the geometrical evolution of a batch in the general random case.

%%%%%%%%%%%%%%%%%%%%%%%%%%%%%%%%%%%%%%%%%%%%%%%%%%%%%%%%%%%%%%%
\subsection*{An Invariant Geometry for Recentering Followed by ReLU}

The following definition is an obvious candidate for an invariant representation under RC+ReLU. It consists of one odd point and a cluster where all points are identical. The cluster and the odd point are orthogonal, and the cluster has a much smaller norm. Such a configuration closely resembles the kind of geometrical structure that we see in practice, see Figure~\ref{fig:geom_ms}.

\begin{definition}[invariant representation]
\label{def:invariant}
    Let $X\tu=\lrp{\left. \bx_1 \right\vert \cdots\middle| \bx_n} \in \reals^{k \times n}$ be a representation of a $d$-dimensional batch of size $n$ 
    % the $d\times n$ matrix representation of a batch 
    at layer $t$, and denote $\bm \nu_{c}\coloneqq \frac{1}{n-1}\sum_{i=2}^n \bm x_i $. We say that $X\tu$ is an invariant representation under RC+ReLU if all the following relations hold.
    \begin{enumerate}
        \item  $\norm{\bm x_1 }^2=1$. 
        \item  $\norm{\bm \nu_{c} }^2=\frac{1}{(n-1)^2}$.
        \item   $  \bx_1  \cdot \bm \nu_{c} = 0  $.
        \item     $\bm x_i =\bm \nu_{c}   $ for all $2 \leq i \leq n$.
    \end{enumerate}
\end{definition}

The following theorem shows that \defref{def:invariant} indeed makes sense: it states that, in expectation, the invariant representation is indeed invariant under an appropriate initialization~variance.

\begin{theorem} \label{thm:invariant}
  Let $X\tu = \lrp{\left. \bx_1 \right\vert \cdots\middle| \bx_n} \in \reals^{k \times n}$ be an 
  % $k\times n$ 
  invariant representation under RC+ReLU, and let $X\tup=\R\lrp{W X\tu -\bm \mu\tu}$ and $\hat{ \bm \nu}_c \coloneqq   {\R \lrp{W \bm \nu_c- \bm \mu\tu}}$, where $\bm\mu\tu = \frac{1}{n}\sum_{i=1}^n W\bx_i\tu$.
 Then,  $\E{X\tup}$ is also an invariant representation under RC+ReLU, where the expectation is over the $d\times k$ matrix $W$ with  i.i.d. $\mathcal{N}(0,\sigma^2)$ entries with $\sigma^2=\frac{2\alpha}{d}$ and $\alpha=\frac{n^2}{n^2-2n+2}$.

\end{theorem}

Finally, the following theorem shows that when starting from a representation somewhat similar to an invariant representation, the next layer brings us closer to an invariant representation in expectation. This is true since items 1--4 in Definition~\ref{def:invariant} correspond to items 1--4 in Theorem~\ref{thm:rand_stb}. As the correspondence between items 1--3 is natural, we focus on item 4. Indeed, after passing through a layer, it is not likely that the cluster would collapse to a single point as in Definition~\ref{def:invariant}. Nonetheless, item 4 shows that the cluster contracts, while keeping the scale of the representation of $\bx_1$ the same.

\begin{theorem} 
\label{thm:rand_stb}
    Let $X\tu=\lrp{\left. \bx_1 \right\vert \cdots\middle| \bx_n} \in \reals^{k \times n}$ be 
    an
    % a $k\times n$ 
    invariant representation such that  $\norm{\bm x_1 - \bm \nu_{c} }=R>\frac{n}{n-1}$, 
    and   $\norm{ \bm x_i -\bm \nu_{c} }  < 1/n^2 $ for all $2 \leq i \leq n$. Then, for 
    $$X\tup = \lrp{\left. \bx\tup_1 \right\vert \cdots\middle| \bx\tup_n} \defeq \R\lrp{W X\tu -\bm \mu\tu}  ~~ \text{and} ~~ \hat{ \bm \nu}_c \coloneqq   {\R \lrp{W \bm \nu_c- \bm \mu\tu}}, $$
    the following relations hold.
    \begin{enumerate}
         \item  $\E{\norm{\bm x\tup_1 }} = 1$.
         \item $\E{\norm{ \hat{ \bm \nu}_c }} =\frac{1}{n-1}$.
         \item $  \bx\tup_1  \cdot\hat{ \bm \nu}_c = 0  $  (This holds for any $W$ and not only in expectation).
         \item Denote $\bx_{n+1}\coloneqq \bm \nu_{c} $ and $\bx\tup_{n+1} \coloneqq \hat{ \bm \nu}_c $. Then, 
         $\E{\norm{ \bx\tup_i - \bx\tup_j     } ^2} <   \norm{  \bx_i - \bx_j   }^2 $  for all $2 \leq i,j  \leq n+1$ such that $\bx\tup_i \neq \bx\tup_j $.
    \end{enumerate}
    where the expectations are with respect to the $d\times k$ matrix $W$ whose entries are i.i.d.\ $\mathcal{N}(0,\sigma^2)$ with $\sigma^2=\frac{2\alpha}{d}$ and $\alpha=\frac{n^2}{(n-1)^2R^2}$.
\end{theorem}

\section*{Discussion}

It remains to be seen how all the different components of BN interact with one another and what are the exact properties that are responsible for the success of BN. Observing the histogram in Figure~\ref{fig:hist_with} and combining it with the insights appearing in this paper and previous work, we explore a new initialization scheme:  The histogram in Figure~\ref{fig:hist_with} implies that the neuron is mostly inactive or with small intensity for most inputs. 

Our work shows that with no rescaling and at initialization, neurons are active with a large intensity only for the odd data point.  Combining this understanding with the insight that orthogonal representations might be beneficial, suggests associating every data point with a unique neuron. That neuron will fire only for its associated input. This way, we attain an orthogonal representation and sparse activity in the network, as we observe in our work. 

Initial experimentation suggests that this might be a good initialization strategy for small data sets, as it would be computationally hard to associate a neuron to every data point since some datasets contain millions of datapoints.

\bibliography{bib}

\begin{thebibliography}{22}
\providecommand{\natexlab}[1]{#1}
\providecommand{\url}[1]{\texttt{#1}}
\expandafter\ifx\csname urlstyle\endcsname\relax
  \providecommand{\doi}[1]{doi: #1}\else
  \providecommand{\doi}{doi: \begingroup \urlstyle{rm}\Url}\fi

\bibitem[Arora et~al.(2019)Arora, Li, and Lyu]{arora2018}
Sanjeev Arora, Zhiyuan Li, and Kaifeng Lyu.
\newblock Theoretical analysis of auto rate-tuning by batch normalization.
\newblock In \emph{International Conference on Learning Representations}, 2019.
\newblock URL \url{https://openreview.net/forum?id=rkxQ-nA9FX}.

\bibitem[Bjorck et~al.(2018)Bjorck, Gomes, Selman, and Weinberger]{bjorck}
Nils Bjorck, Carla~P Gomes, Bart Selman, and Kilian~Q Weinberger.
\newblock Understanding batch normalization.
\newblock In S.~Bengio, H.~Wallach, H.~Larochelle, K.~Grauman, N.~Cesa-Bianchi, and R.~Garnett, editors, \emph{Advances in Neural Information Processing Systems}, volume~31. Curran Associates, Inc., 2018.
\newblock URL \url{https://proceedings.neurips.cc/paper/2018/file/36072923bfc3cf47745d704feb489480-Paper.pdf}.

\bibitem[Cho and Saul(2009)]{cho}
Youngmin Cho and Lawrence Saul.
\newblock Kernel methods for deep learning.
\newblock In Y.~Bengio, D.~Schuurmans, J.~Lafferty, C.~Williams, and A.~Culotta, editors, \emph{Advances in Neural Information Processing Systems}, volume~22. Curran Associates, Inc., 2009.
\newblock URL \url{https://proceedings.neurips.cc/paper/2009/file/5751ec3e9a4feab575962e78e006250d-Paper.pdf}.

\bibitem[Daneshmand et~al.(2020)Daneshmand, Kohler, Bach, Hofmann, and Lucchi]{bach20}
Hadi Daneshmand, Jonas Kohler, Francis Bach, Thomas Hofmann, and Aurelien Lucchi.
\newblock Batch normalization provably avoids ranks collapse for randomly initialised deep networks.
\newblock In H.~Larochelle, M.~Ranzato, R.~Hadsell, M.F. Balcan, and H.~Lin, editors, \emph{Advances in Neural Information Processing Systems}, volume~33, page 18387–18398. Curran Associates, Inc., 2020.
\newblock URL \url{https://proceedings.neurips.cc/paper/2020/file/d5ade38a2c9f6f073d69e1bc6b6e64c1-Paper.pdf}.

\bibitem[Daneshmand et~al.(2021)Daneshmand, Joudaki, and Bach]{bach21}
Hadi Daneshmand, Amir Joudaki, and Francis Bach.
\newblock Batch normalization orthogonalizes representations in deep random networks.
\newblock In M.~Ranzato, A.~Beygelzimer, Y.~Dauphin, P.S. Liang, and J.~Wortman Vaughan, editors, \emph{Advances in Neural Information Processing Systems}, volume~34, page 4896–4906. Curran Associates, Inc., 2021.
\newblock URL \url{https://proceedings.neurips.cc/paper/2021/file/26cd8ecadce0d4efd6cc8a8725cbd1f8-Paper.pdf}.

\bibitem[De and Smith(2020)]{de2020batch}
Soham De and Sam Smith.
\newblock Batch normalization biases residual blocks towards the identity function in deep networks.
\newblock \emph{Advances in Neural Information Processing Systems}, 33:\penalty0 19964--19975, 2020.

\bibitem[Dittmer et~al.(2020)Dittmer, King, and Maass]{sv_NN}
Sören Dittmer, Emily~J. King, and Peter Maass.
\newblock Singular values for relu layers.
\newblock \emph{IEEE Transactions on Neural Networks and Learning Systems}, 31\penalty0 (9):\penalty0 3594--3605, 2020.
\newblock \doi{10.1109/TNNLS.2019.2945113}.

\bibitem[Frankle et~al.(2020)Frankle, Schwab, and Morcos]{frankle2020training}
Jonathan Frankle, David~J Schwab, and Ari~S Morcos.
\newblock Training batchnorm and only batchnorm: On the expressive power of random features in cnns.
\newblock \emph{arXiv preprint arXiv:2003.00152}, 2020.

\bibitem[Furusho and Ikeda(2020)]{furusho2020theoretical}
Yasutaka Furusho and Kazushi Ikeda.
\newblock Theoretical analysis of skip connections and batch normalization from generalization and optimization perspectives.
\newblock \emph{APSIPA Transactions on Signal and Information Processing}, 9:\penalty0 e9, 2020.

\bibitem[He et~al.(2015)He, Zhang, Ren, and Sun]{kaiming}
Kaiming He, Xiangyu Zhang, Shaoqing Ren, and Jian Sun.
\newblock Delving deep into rectifiers: Surpassing human-level performance on imagenet classification.
\newblock \emph{CoRR}, abs/1502.01852, 2015.
\newblock URL \url{http://arxiv.org/abs/1502.01852}.

\bibitem[He et~al.(2016)He, Zhang, Ren, and Sun]{he}
Kaiming He, Xiangyu Zhang, Shaoqing Ren, and Jian Sun.
\newblock Deep residual learning for image recognition.
\newblock In \emph{2016 IEEE Conference on Computer Vision and Pattern Recognition (CVPR)}, pages 770--778, 2016.
\newblock \doi{10.1109/CVPR.2016.90}.

\bibitem[Huang et~al.(2017)Huang, Liu, Van Der~Maaten, and Weinberger]{huang}
Gao Huang, Zhuang Liu, Laurens Van Der~Maaten, and Kilian~Q. Weinberger.
\newblock Densely connected convolutional networks.
\newblock In \emph{2017 IEEE Conference on Computer Vision and Pattern Recognition (CVPR)}, pages 2261--2269, 2017.
\newblock \doi{10.1109/CVPR.2017.243}.

\bibitem[Ioffe and Szegedy(2015)]{bn}
Sergey Ioffe and Christian Szegedy.
\newblock Batch normalization: Accelerating deep network training by reducing internal covariate shift.
\newblock In \emph{International conference on machine learning}, pages 448--456. PMLR, 2015.

\bibitem[Karakida et~al.(2019)Karakida, Akaho, and Amari]{karakida}
Ryo Karakida, Shotaro Akaho, and Shun-ichi Amari.
\newblock The normalization method for alleviating pathological sharpness in wide neural networks.
\newblock In H.~Wallach, H.~Larochelle, A.~Beygelzimer, F.~d\textquotesingle Alch\'{e}-Buc, E.~Fox, and R.~Garnett, editors, \emph{Advances in Neural Information Processing Systems}, volume~32. Curran Associates, Inc., 2019.
\newblock URL \url{https://proceedings.neurips.cc/paper/2019/file/9edda0fd4d983bf975935cfd492fd50b-Paper.pdf}.

\bibitem[Li et~al.(2019)Li, Chen, Hu, and Yang]{li2019understanding}
Xiang Li, Shuo Chen, Xiaolin Hu, and Jian Yang.
\newblock Understanding the disharmony between dropout and batch normalization by variance shift.
\newblock In \emph{Proceedings of the IEEE/CVF conference on computer vision and pattern recognition}, pages 2682--2690, 2019.

\bibitem[Lubana et~al.(2021)Lubana, Dick, and Tanaka]{lubana2021beyond}
Ekdeep~S Lubana, Robert Dick, and Hidenori Tanaka.
\newblock Beyond batchnorm: Towards a unified understanding of normalization in deep learning.
\newblock \emph{Advances in Neural Information Processing Systems}, 34:\penalty0 4778--4791, 2021.

\bibitem[Lyu et~al.(2022)Lyu, Li, and Arora]{lyu2022understanding}
Kaifeng Lyu, Zhiyuan Li, and Sanjeev Arora.
\newblock Understanding the generalization benefit of normalization layers: Sharpness reduction.
\newblock \emph{arXiv preprint arXiv:2206.07085}, 2022.

\bibitem[Santurkar et~al.(2018)Santurkar, Tsipras, Ilyas, and Madry]{santukar}
Shibani Santurkar, Dimitris Tsipras, Andrew Ilyas, and Aleksander Madry.
\newblock How does batch normalization help optimization?
\newblock In S.~Bengio, H.~Wallach, H.~Larochelle, K.~Grauman, N.~Cesa-Bianchi, and R.~Garnett, editors, \emph{Advances in Neural Information Processing Systems}, volume~31. Curran Associates, Inc., 2018.
\newblock URL \url{https://proceedings.neurips.cc/paper/2018/file/905056c1ac1dad141560467e0a99e1cf-Paper.pdf}.

\bibitem[Saxe et~al.(2014)Saxe, Mcclelland, and Ganguli]{Saxe14exactsolutions}
Andrew~M. Saxe, James~L. Mcclelland, and Surya Ganguli.
\newblock Exact solutions to the nonlinear dynamics of learning in deep linear neural network.
\newblock In \emph{In International Conference on Learning Representations}, 2014.

\bibitem[Silver et~al.(2017)Silver, Schrittwieser, Simonyan, Antonoglou, Huang, Guez, Hubert, Baker, Lai, Bolton, Chen, Lillicrap, Hui, Sifre, van~den Driessche, Graepel, and Hassabis]{silver}
David Silver, Julian Schrittwieser, Karen Simonyan, Ioannis Antonoglou, Aja Huang, Arthur Guez, Thomas Hubert, Lucas Baker, Matthew Lai, Adrian Bolton, Yutian Chen, Timothy Lillicrap, Fan Hui, Laurent Sifre, George van~den Driessche, Thore Graepel, and Demis Hassabis.
\newblock Mastering the game of go without human knowledge.
\newblock \emph{Nature}, 550:\penalty0 354--, October 2017.
\newblock URL \url{http://dx.doi.org/10.1038/nature24270}.

\bibitem[Wang et~al.(2022)Wang, Wu, and Huang]{wang2022understanding}
Jiaxi Wang, Ji~Wu, and Lei Huang.
\newblock Understanding the failure of batch normalization for transformers in nlp.
\newblock \emph{arXiv preprint arXiv:2210.05153}, 2022.

\bibitem[Yang et~al.(2019)Yang, Pennington, Rao, Sohl-Dickstein, and Schoenholz]{yang2019mean}
Greg Yang, Jeffrey Pennington, Vinay Rao, Jascha Sohl-Dickstein, and Samuel~S Schoenholz.
\newblock A mean field theory of batch normalization.
\newblock \emph{arXiv preprint arXiv:1902.08129}, 2019.

\end{thebibliography}
\clearpage
\appendix

\section{Figures}

\begin{figure}[h!]
\centering

\subfigure[BN without RC (=NL+RS)][t]{
    \includegraphics[trim=0 500 0 500 , clip, width=0.48\linewidth]{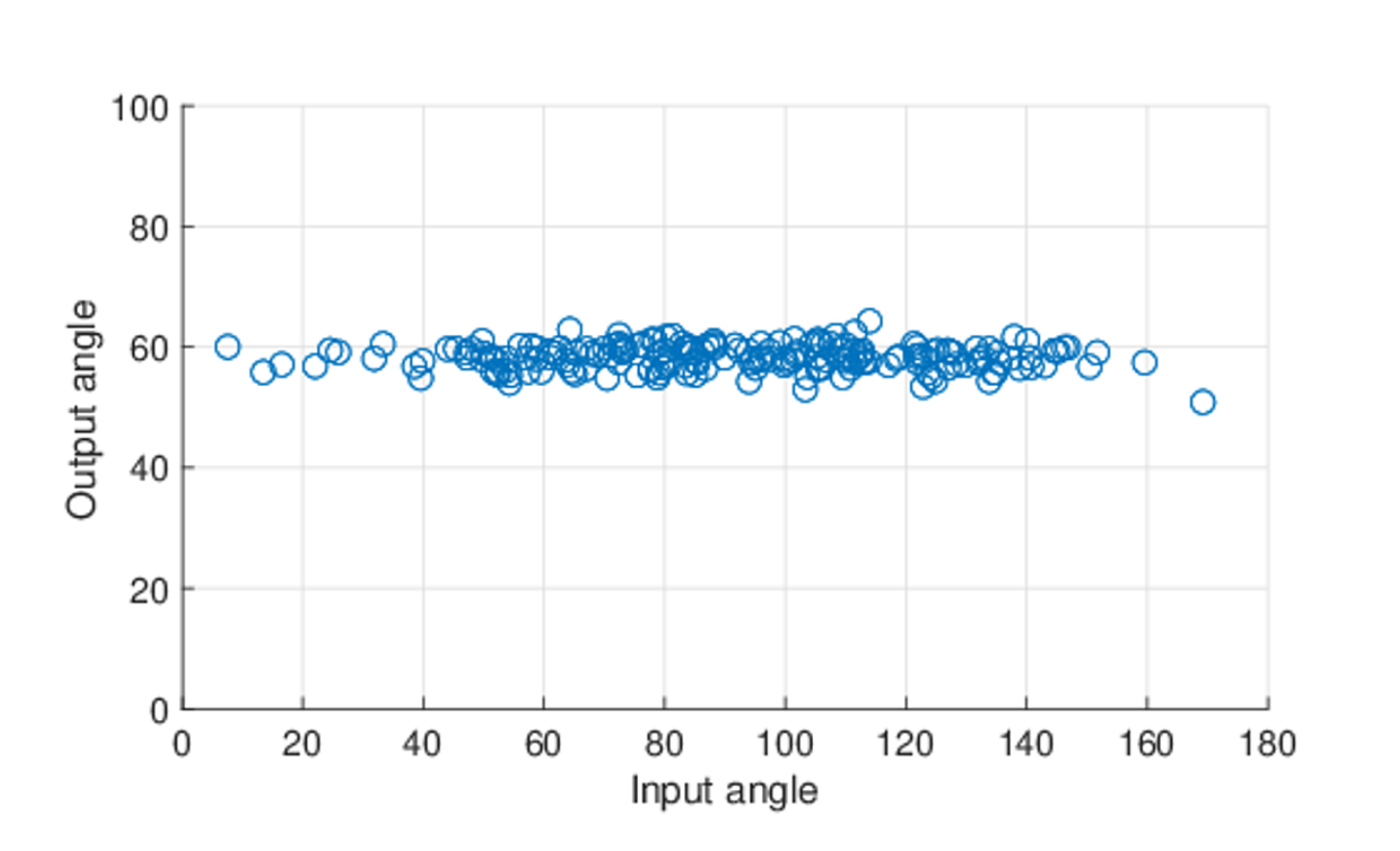}
    \label{fig:angle_no_mean}
} 
\subfigure[BN (=NL+RC+RS) ][t]{
    \includegraphics[trim=0 500 0 500 , clip, width=0.48\linewidth]{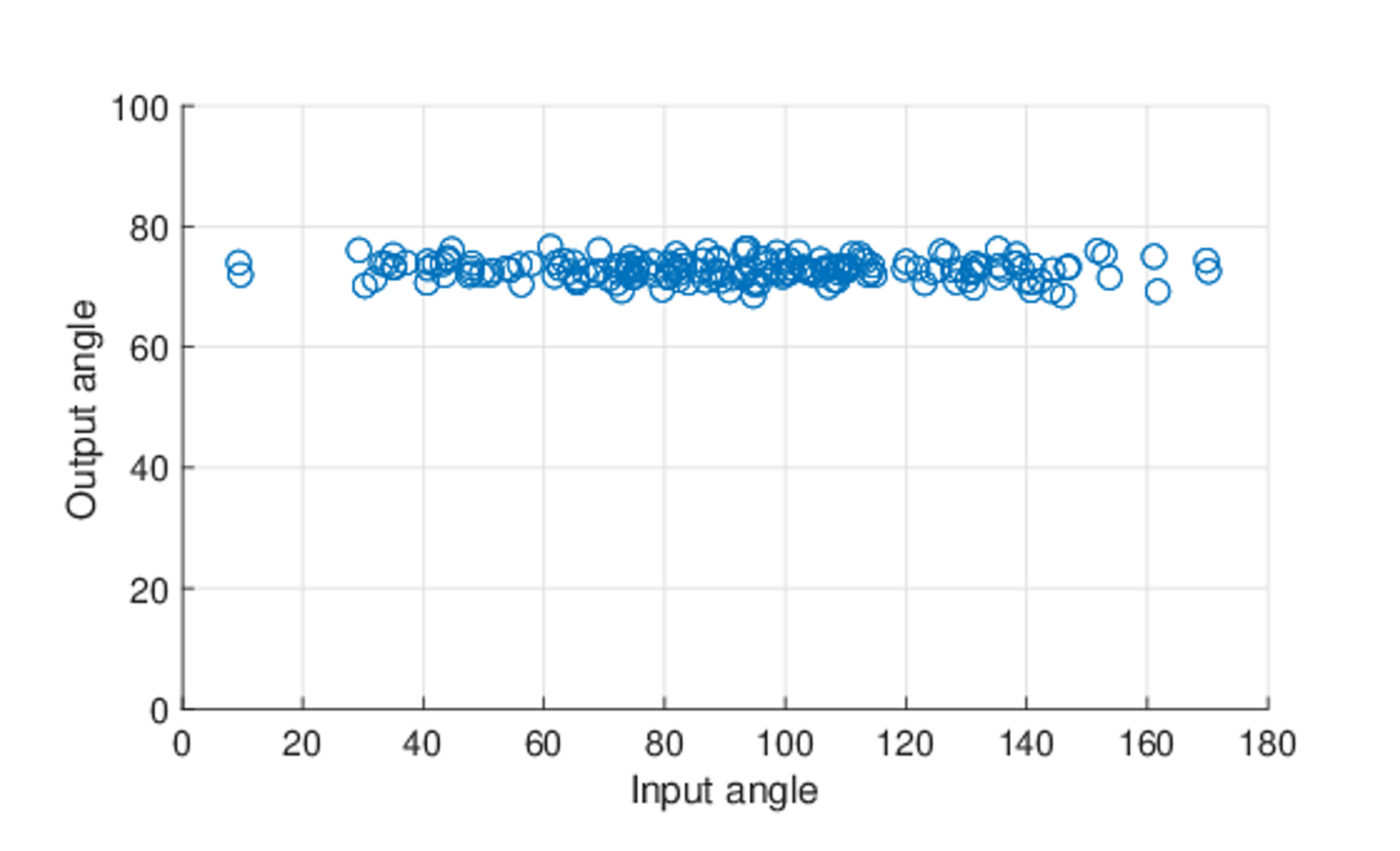}
    \label{fig:angle_mean}
} % END \subfigure

  \caption{Effect of recentering on the angles between pairs of data points before and after the 30th layer of the neural network. If Batch Normalization without recentering is used, the output angles are approximately $60^{\circ}$. With recentering, the angles increase to approximately $75^{\circ}$.}
\label{fig:no_mean}

\end{figure}

\begin{figure}[h!]
\centering
\subfigure[Without RC][t]{ 
        \includegraphics[trim=0 400 0 500 , clip,width=0.45\linewidth]{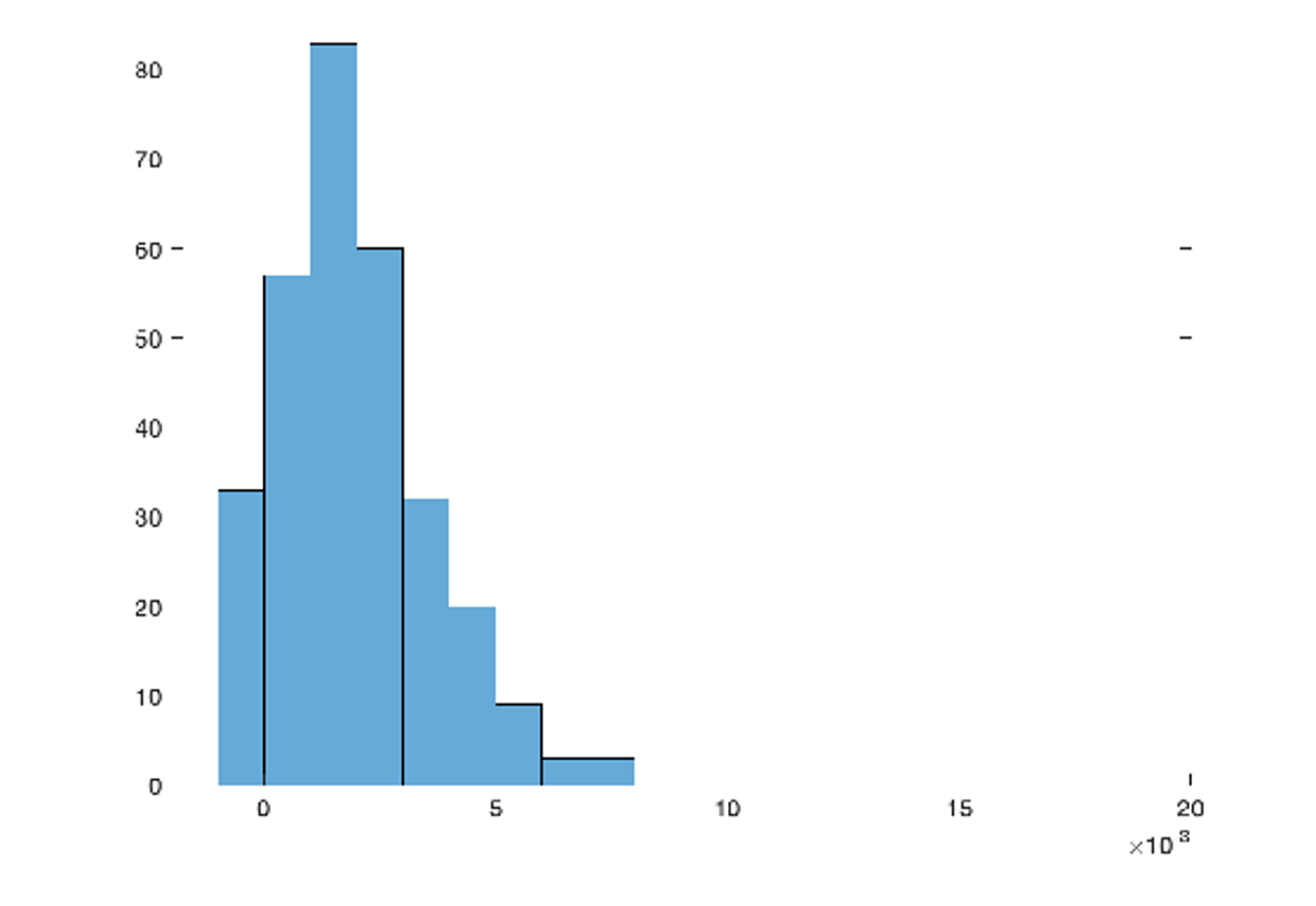}
        \label{fig:hist_no}
    }
  \subfigure[With RC][t]{ 
        \includegraphics[trim=0 400 0 500 , clip,width=0.45\linewidth]{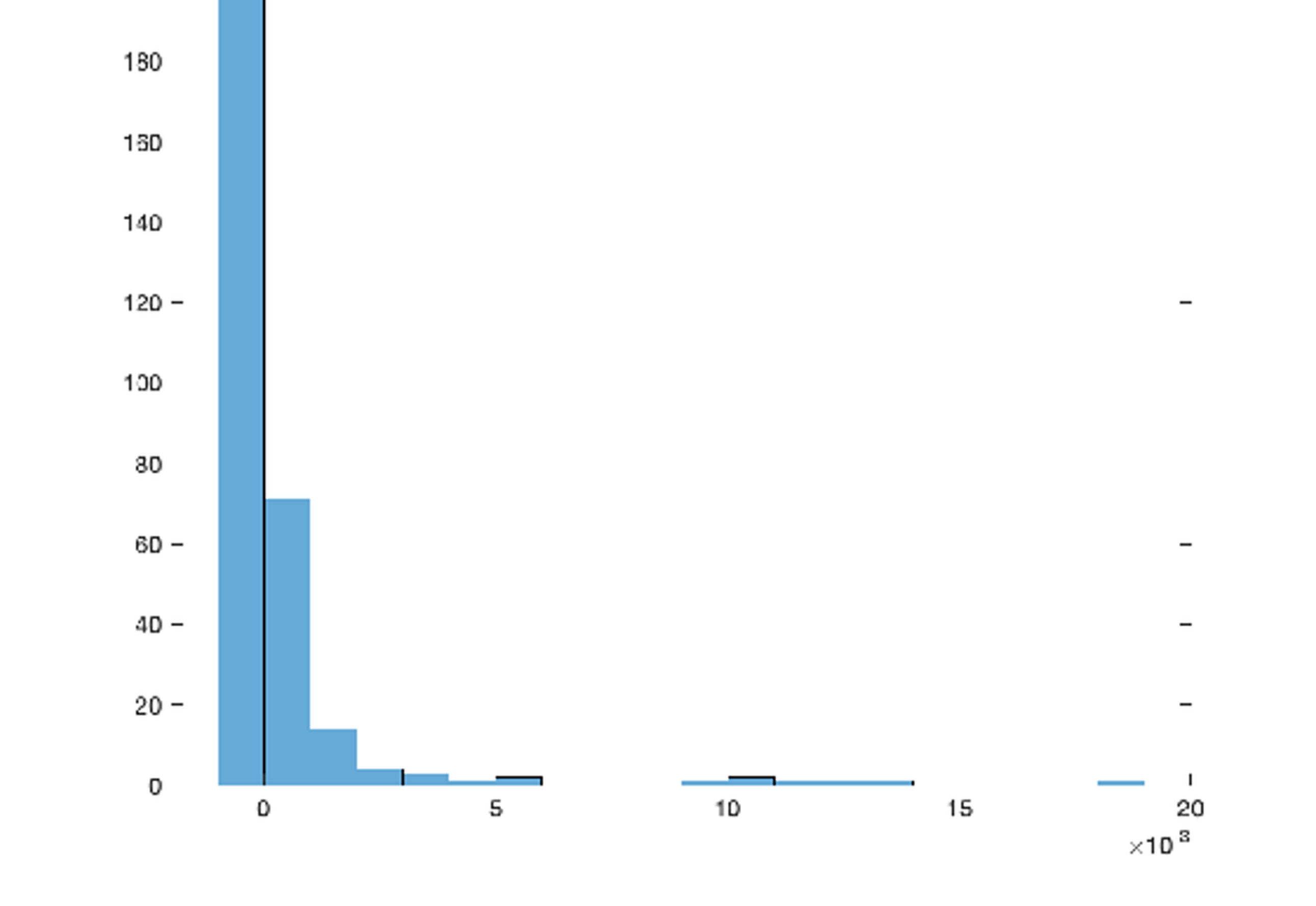}
        \label{fig:hist_with}
    }
 \caption{Histograms that represent the activity for a neuron in a network with and without RC. Without RC, some neurons exhibit neural activity like the one in Figure (a): the neuron has significant response for a large number of inputs. With RC, most of the neurons behave according to Figure (b): the neural response is negligible for most of the inputs.}
\label{fig:hist}

\end{figure}

\section{Proof of Theorem~\ref{thm:rank} }

\begin{proof}[Lemma \ref{lem:rank}]
    % Denote the matrix composed of the first $\ell$ \emph{columns} of $X$ by $X^{(\ell)}$, 
    % and the matrix composed of the first $\ell$ \emph{rows} of $W$ by $W^{(\ell)}$.
    For $\rank \lrp{ \R \lrp{ W^{(r)} X^{(t)} } } = t$, the relation
    \begin{align}
        \sum_{j=1}^t a_j \R \lrp{W^{(r)} \bx_j} = \R \lrp{W^{(r)} \bx_{t+1}} 
    \end{align}
    either does not hold, in which case $\rank \lrp{ \R \lrp{ W^{(r+1)} X^{(t+1)} } } = t+1$ holds with probability 1 and \eqref{eq:lem:rank} holds, or the relation holds for a \emph{unique} set of $\{a_i\}$. Therefore, we are left with treating the latter case. 
    In that case, for the rank to remain $t$, 
    \begin{subequations}
    \begin{align}
    \label{eq:rank-remains:W^r}
        \sum_{j=1}^{t+1} a_j \R \lrp{W^{(r)} \bx_j} = 0
        % \R \lrp{ W_{t+1} X_{t+1} }
    \end{align}
    has to hold with $a_{t+1} \defeq -1$. Now, for \eqref{eq:lem:rank} \textit{not} to hold, 
    \begin{align}
    \label{eq:rank-remains:last_w}
        \sum_{j=1}^{t+1} a_j \R \lrp{\rw_{r+1} \bx_j} = 0 
        % \R \lrp{ W_{t+1} X_{t+1} }
    \end{align}
    has to hold with the same coefficients $\{a_j\}$ as in \eqref{eq:rank-remains:W^r}, namely, 
    \begin{align}
    \label{eq:rank-remains:W^r+1}
        \sum_{j=1}^{t+1} a_j \R \lrp{W^{(r+1)} \bx_j} = 0 .
        % \R \lrp{ W_{t+1} X_{t+1} }
    \end{align}
    \end{subequations}

    We next prove that \eqref{eq:rank-remains:W^r+1} does not hold if, 
    for some $\ell \in [r]$ and $i \in [t+1]$, 
    \begin{subequations}
    \label{eq:one-ineq}
    \noeqref{eq:one-ineq:ineq,eq:one-ineq:eqs}
    \begin{align}
        % &\exists i \in [t+1]: 
        \sign \lrp{ \rw_{r+1} \bx_i } &\neq \sign \lrp{ \rw_\ell \bx_i }, 
        & a_i \neq 0 ,
    \label{eq:one-ineq:ineq}
     \\ \sign \lrp{ \rw_{r+1} \bx_j } &= \sign \lrp{ \rw_\ell \bx_j } 
        & \forall j \neq i .
    \label{eq:one-ineq:eqs}
    \end{align}
    \end{subequations} 
    Furthermore, \eqref{eq:one-ineq} for $\ell \in [r]$, such that $\rw_\ell \neq 0$, holds with probability of at least $\gamma \lrp{ X^{(t+1)} }$.

    % Assume that \eqref{eq:one-ineq} holds for row $\ell \in [r]$ and that 
    Take some $\ell \in [r]$ for which $\rw_\ell \neq 0$ (there must exist such a row by the lemma assumption),
    and denote $\cJ_\ell \defeq \lrbrace{ j \in [t+1] :\ \rw_\ell \bx_j > 0, a_j \neq 0}$. Then, according to \eqref{eq:rank-remains:W^r}, 
    \begin{align}
    \label{eq:X_orthog_W}
        \sum_{j=1}^{t+1} a_j \R \lrp{\rw_\ell \bx_j}
        =  \rw_\ell \sum_{j \in \cJ_\ell} a_j \bx_j = \rw_\ell \bu = 0 ,
    \end{align}
    with $\bu \defeq \sum_{j \in \cJ} a_j \bx_j$, namely 
    $\rw_\ell \perp \bu^T$.

    If $\bu \neq 0$, then $\rw_{r+1} \bu \neq 0$ for $\rw_{r+1}$ in the equivalence class of $\rw_\ell$ (recall \defref{eq:equivalence}) up to measure zero, namely, with probability of at least 
    $\gamma \lrp{\tilde{X}^{(t+1)}}$, where the columns of $\tilde{X}^{(t+1)}$ are all the vectors in the set $\lrbrace{ \bx_j : a_j \neq 0, j \in [t+1] }$; clearly, $\gamma \lrp{\tilde{X}^{(t+1)}} \geq \gamma \lrp{X^{(t+1)}}$.
    
    If $\bu = 0$, on the other hand, then

   \begin{align}
    \label{eq:X_non-orthog_W}
        \sum_{j=1}^{t+1} a_j \R \lrp{\rw_{r+1} \bx_j}
        =  \rw_{r+1} \lrp{\bu \pm a_i \bx_i} = \pm a_i \rw_{r+1} \bx_i \neq 0 
    \end{align}
    for $\rw_{r+1}$ that satisfies \eqref{eq:one-ineq};
    we will show next that such $\rw_{r+1}$ constitute an equivalence class, 
    meaning that \eqref{eq:X_non-orthog_W} holds, again, with probability of at least $\gamma \lrp{X^{(t+1)}}$.

    We are left with proving that the set of $\rw_{r+1}$ that satisfy \eqref{eq:one-ineq} for a given $\ell$ for which $\rw_\ell \neq 0$, and some $i$, 
    consitutues an equivalence class of non-zero volume.
    To that end, denote 
    \begin{align} 
        i &\defeq \argmin_{i \in [t+1]: a_i \neq 0} \abs{\rw_\ell \bx_i} .
        % & p &\defeq \abs{\rw_\ell \bx_i}, 
    \end{align} 
    Then, for a sufficiently small $\eps > 0$, 
    \begin{align} 
        \lrp{ \rw_\ell - \lrp{1 + \eps} \frac{\rw_\ell \bx_i}{\norm{\bx_i} } \bx_i^T } \bx_i &< 0
     \\ \lrp{ \rw_\ell - \lrp{1 + \eps} \frac{\rw_\ell \bx_i}{\norm{\bx_i} } \bx_i^T } \bx_i &> 0
        & \forall j \in \lrbrace{ j \in [t+1] :\ j \neq i, a_j \neq 0} ,
    \end{align} 
    since there are now collinear vectors in $\lrbrace{\bx_j :\ j \in [t+1]}$.
\end{proof}

\begin{proof}[Theorem \ref{thm:rank}]
Denote the matrix composed of the first $\ell$ \emph{columns} of $X$ by $X^{(\ell)}$.
% and the matrix composed of the first $\ell$ \emph{rows} of $W$ by $W^{(\ell)}$.
% $X=\lrp{x_1|...|x_n}$,  $X^{ \lrp{t} }= \lrp{x_1|...|x_t} $, $W=\lrp{w_1 ;  ...; w_d}$, and $ W^{\lrp{r}}=\lrp{w_1 ;  ...; w_r}  $.
Denote further $r_t \defeq \min\lrbrace{ r \in \nats:\ \R \lrp{W^{(r)} X^{(r)}} = t }$ and 
$y_t \defeq r_t - r_{t-1}$ with $r_0 \defeq 0$.
Clearly, $y = \sum_{t=1}^n y_t$.

By the definition of $r_t$, 
\begin{align} 
    \rank \lrp{\R \lrp{ W^{(r_t)} X^{(t)} }} = t .
\end{align} 
Then, by \lemref{lem:rank},
\begin{align} 
    \rank \lrp{\R \lrp{ W^{(r_t+1)} X^{(t+1)} }} = t+1
\end{align} 
with probability of at least $\gamma \lrp{X^{(t+1)}}$. 
If 
\begin{align} 
    \rank \lrp{\R \lrp{ W^{(r_t+1)} X^{(t+1)} }} = t, 
\end{align} 
add another row; then, 
\begin{align} 
    \rank \lrp{\R \lrp{ W^{(r_t+2)} X^{(t+1)} }} = t+1
\end{align} 
with probability of at least $\gamma \lrp{X^{(t+1)}}$. Continue adding rows until the first $r$ for which 
\begin{align} 
    \rank \lrp{\R \lrp{ W^{(r_t+r)} X^{(t+1)} }} = t+1. 
\end{align} 
Identifying that $y_t = r$ and noting that the cumulative distribution function (CDF) of $r$ is majorized by that of a geometric distribution with success probability $\gamma \lrp{X^{(t+1)}}$, yields 
\begin{align} 
    \E{y_t} \leq \frac{1}{\gamma \lrp{X^{(t+1)}}} .
\end{align} 
By noting that $\gamma \lrp{ X\tu }$ is a monotonically decreasing function in $t$, yields
\begin{align}
    \E{Y} = \sum_{t=1}^n \E{y_t} \leq \sum_{t=1}^n \frac{1}{\gamma \lrp{X^{(t+1)}}} \leq \frac{n}{\gamma} \,.
\end{align}
This concludes the proof of item 1.

To prove item 2, note first that $y_1, y_2, \ldots, y_n$ are independent.
Then, 

 \begin{subequations}
    \label{eq:proof:rank:whp}
    \begin{align} 
        &\P{\rank(\R(W^{(d)} X)) < n}
        \leq \P{  y_1+\cdots+y_n  >  \frac{\alpha n}{\gamma} }
    \label{eq:proof:rank:whp:iid-geometric}
     \\ &\qquad\qquad \leq \min_{s > 0} \left( \frac{\gamma \Exp{s}}{1 - (1-\gamma) \Exp{s}} \right)^n \Exp{-s \frac{\alpha n}{\gamma}}
    \label{eq:proof:rank:whp:chernoff}
     % \\ &\qquad\qquad = \Exp{n \left( \log \frac{\alpha - \gamma}{1 - \gamma} + \frac{\alpha}{\gamma} \left( \log(1-\gamma) - \log \frac{\alpha - \gamma}{\alpha} \right) \right)}
     \\ &\qquad\qquad = \Exp{-n \lrp{ \frac{\alpha}{\gamma} \log \frac{1}{\alpha} + \lrp{\frac{\alpha}{\gamma} - 1} \log \frac{\alpha - \gamma}{1 - \gamma} }}
    \label{eq:proof:rank:whp:chernoff:explicit}
     \\ &\qquad\qquad \leq \Exp{-n \lrp{ \alpha - 1 - \log \alpha }}
    \label{eq:proof:rank:whp:gamma-->0}
    %  (1-\gamma ) ^{i-n}  \gamma ^n
    % \label{eq:proof:rank:whp:binomial}
% \\& \leq  2  {  \frac { \alpha n} {\gamma}  -1 \choose  n-1 }
%      (1-\gamma ) ^{ \frac { \alpha n} {\gamma}-n}  \gamma ^n
% \\& \leq  2  {  \frac { \alpha n} {\gamma}  \choose  n }
%      (1-\gamma ) ^{ \frac { \alpha n} {\gamma}-n}  \gamma ^n
%  \\&    \leq    \lrp{\frac{\alpha  e}{\gamma }}^n  \gamma ^n    (1-\gamma ) ^{ \frac { \alpha n} {\gamma}-n}
%  \\  &    \leq    \lrp{{\alpha e}}^n \exp\lrp{-\gamma n \lrp{ ֿ\frac{\alpha }{\gamma} -1   }  } 
% \\  &    \leq   \alpha^n \exp\lrp{-\gamma n \lrp{\frac{\alpha -2}{\gamma}    }  } 
% \\ & \stackrel{?}{\approx}   \alpha^n \exp\lrp{-n \lrp{ \alpha -2   }  } 
 \end{align}
\end{subequations}
    where 
    \eqref{eq:proof:rank:whp:iid-geometric} follows from the definition of $\{y_t\}$;
    \eqref{eq:proof:rank:whp:chernoff} follows from Chernoff's inequality by recalling that $y_1, \ldots, y_n$ are independent and that their (marginal) CDFs are majorized by that of a geometric distribution with success probability $\gamma$;
    and \eqref{eq:proof:rank:whp:gamma-->0} follows from the monotonicity in $\gamma$ of the expression in \eqref{eq:proof:rank:whp:chernoff:explicit}.

\end{proof}

\section{Proof of Theorem~\ref{ThmStable}}

    By \lemref{lem:c(f(x))<=c(x)}, the sequence $ \left\{ c\left( \rx^{(t)} \right) : t \in \nats \cup \{0\} \right\}$ is non-increasing.    
    Assume that $\rx^{(0)}$ is not stable, i.e., $c\left( \rx^{(0)} \right) > 3$, %it is composed of more than 3 clusters 
    as otherwise the result is trivially true. 
% Notice that 
    % By \lemref{lem:c(f(x))<=c(x)}, 
    It suffices to prove that there exists a finite $\tau \in \nats$ such that $\rx^{(\tau)}$ is more clustered than $\rx^{(0)}$. We start by proving this result in the case where the infinite sequence of transformations involves only positive transformations. 
    Assume by contradiction that $\rx^{(t)}$ is not more clustered than $\rx^{(0)}$ for all $t \in \nats$, i.e., %$c\left(\bx^{(t)}\right) = c\left(\bx^{(0)}\right)$.
    % Since positive and negative transformations cannot make a vector less stable (in fact, no deterministic transform can make a vector less stable), this means that
    % Then, %by \lemref{lem:c(f(x))<=c(x)}, 
    \begin{align}
    \label{eq:CAssumption}
        c \left( \rx^{(t)} \right) &= c \left( \rx^{(0)} \right) > 3 & \forall t \in \nats.
    \end{align}
    Let $n_t \defeq \left\lvert \left\{ i: x^{(t)}_i > 0 \right\} \right\rvert$, and define
    \begin{align}
    \Delta_t \defeq c_3 \left(\rx^{(t)} \right) - c_2 \left( \rx^{(t)} \right).
    \end{align}
    By the assumption in \eqref{eq:CAssumption}, 
    % \MG{Detail: I don't like this eqref format. I much prefer Equation (4). You?}, 
    % \begin{subequations} 
    \begin{align}
    \label{eq:Delta>0_nt>=3}
        \Delta_{t_1} &= \Delta_{t_2} := \Delta > 0,
      & n_{t_1} &= n_{t_2} \geq 2, 
    \end{align}
    % \end{subequations}
    for all $t_1, t_2 \in \nats$. Thus, 
    since, for all $t \in \nats$, $x_i^{(t)} \geq 0$ for all $i$, we have
    \begin{align}
    \label{MeanBound}
        \ox^{(t)} &= \frac{1}{n}\sum_{i=1}^n x^{(t)}_i > \frac{\Delta}{n} & \forall t \in \nats .
    \end{align}
    
    On the other hand, 
    % by \eqref{eq:CAssumption}, 
    all the strictly positive entries $\bx^{(t)}$ (which are $n_t$) are all greater than the mean $\ox^{(t)}$, as otherwise they would all go to zero at $t+1$, in contradiction to \eqref{eq:CAssumption}.
    Therefore, 
    \begin{align}
    \label{eq:pos-transform-only}
        \ox^{(t+1)} 
        = \frac{1}{n}\sum_{i=1}^n x_i^{(t+1)} 
        = \frac{1}{n}\sum_{i=1}^n \R \left( x^{(t)}_i - \ox^{(t)} \right) 
        = \ox^{(t)} - \frac{n_t}{n}\ox^{(t)} 
        = \left(1 - \frac{n_t}{n}\right)\ox^{(t)},
    \end{align}
    from which it follows that $\lim_{t\to\infty} \ox^{(t)} = 0$ since $n_t \geq 2$ for all $t \in \nats$. This contradicts \eqref{MeanBound}, thus proving the result.
    
    The case of a \textit{finite} number of negative transformations simply follows from the previous case, by repeating the previous proof from layer $t \in \nats$ that corresponds to the last negative transformation in lieu of layer 0.
    % \footnote{This proof implies that, for a finite number of negative transformations, the number of clusters will converge to at most~2.}
    
    Finally, consider the case where there are an infinite number of negative transformations, and assume again \eqref{eq:CAssumption} by contradiction.
    % that $c(\bx^{(t)}) = c(\bx^{(0)})$ for every $t\geq 1$. 
    Consider any two steps $t_1$ and $t_2$ of the sequence such that a negative transformation occurs at $t_1$ and $t_2$, and only positive transformations occur between these two steps,
    viz.
    % To be formal, suppose that
    \begin{subequations}
    \label{eq:pos-between-neg}
    \begin{align}
        \rx^{(t_1)} &= \R(-\rx^{(t_1-1)} + \ox^{(t_1-1)}),
    \label{eq:pos-between-neg:first}
    \\ \rx^{(t)} &= \R(\rx^{(t-1)} - \ox^{(t-1)}), & t_1 < t < t_2 .
    \label{eq:pos-between-neg:inter}
    \\ \rx^{(t_2)} &= \R(-\rx^{(t_2-1)} + \ox^{(t_2-1)}),
    \label{eq:pos-between-neg:last}
    \end{align}
    \end{subequations}
    % and for any eventual $t_1 < t < t_2$,
    % \begin{align}
    % \bx^{(t)} = \R(\bx^{(t-1)} - \ox^{(t-1)}).
    % \end{align}
    We now show that this sequence of transformations decreases the mean, i.e., $\ox^{(t_2)} < \ox^{(t_1-1)}$.
    To that end, denote %the maximal value in $\bx^{(t)}$ by
    $x_{\max}^{(t)} \defeq \max \rx^{(t)}$, 
    and $I_t \defeq \{i: x^{(t)}_i < x_{\max}^{(t)}\}$.
    % $C_{\max}^{(t)} = \max C(\bx^{(t)})$, 
    % $I_t = \{i: x^{(t)}_i < c_{\max}^{(t)}\}$. %, and ${n'' = \lvert\{i: x_i^{(t_2)} > 0\}\rvert}$. 
    Then, %we have
    \begin{subequations}
    \label{eq:inter-pos-transforms}
    \begin{align}
        \ox^{(t_2)} 
        &= \frac{1}{n}\sum_{i=1}^n x_i^{(t_2)} 
    \label{eq:inter-pos-transforms:mean-def}
     \\ &
        = \frac{1}{n}\sum_{i\in I_{t_2-1}} \left( -x_i^{(t_2-1)} + \ox^{(t_2-1)} \right) 
    \label{eq:inter-pos-transforms:assumption}
     \\ &
        = \left(1-\frac{n_{t_2}}{n}\right) \left( x_{\max}^{(t_2-1)} - \ox^{(t_2-1)} \right) 
    \label{eq:inter-pos-transforms:x-max}
     \\ &
        < \left(1-\frac{n_{t_2}}{n}\right) \left( x_{\max}^{(t_1)} - \ox^{(t_1)} \right) 
    \label{eq:inter-pos-transforms:pos-transform}
     \\ &
        % = \left(1-\frac{n_{t_2}}{n}\right) \ox^{(t_1-1)} 
        = \left(1-\frac{n_{t_2}}{n}\right) \left( \ox^{(t_1-1)} - \ox^{(t_1)} \right)
    \label{eq:inter-pos-transforms:neg-to-pos}
     \\ &
        < \left( 1 - \frac{2}{n} \right) \ox^{(t_1-1)} 
    \label{eq:inter-pos-transforms:contraction}
    \end{align}
    \end{subequations}
    where 
    \eqref{eq:inter-pos-transforms:assumption} follows from \eqref{eq:pos-between-neg:last} and 
    the assumption in \eqref{eq:CAssumption}, 
    which suggests that exactly one of the clusters will be nullified by the ReLU activation, as otherwise $c\left( \rx^{(t_2-1)} \right) = 1$ or 
    $c \left( \rx^{(t_2)} \right) < c \left( \rx^{(t_2-1)} \right)$, in contradiction to \eqref{eq:CAssumption};
    \eqref{eq:inter-pos-transforms:x-max} follows from the definition of $n_{t_2}$, $x_{\max}^{(t_2-1)}$, and $\ox^{(t_2-1)}$;
    \eqref{eq:inter-pos-transforms:pos-transform} follows from \eqref{eq:pos-transform-only};
    \eqref{eq:inter-pos-transforms:neg-to-pos} follows from \eqref{eq:pos-between-neg:first} by noting that $x_{\max}^{(t_1)} = \ox^{(t_1-1)}$; 
    and \eqref{eq:inter-pos-transforms:contraction} holds since $\ox^{(t_1)} > 0$ and $n_{t_2} \geq 2$ by \eqref{eq:Delta>0_nt>=3}.
    
    % the last inequality follows because $x_{\max}^{(t_1)} = \ox^{(t_1-1)}$. 
    Hence, a sequence of negative transformations, with each pair of consecutive such transformations possibly separated by several intermediate positive transformations, contracts the vector mean, meaning that $\lim_{t\to\infty} \ox^{(t)} = 0$. 
    However, since \eqref{MeanBound} still holds for this case, we again reach a contradiction, and the theorem is proved.

\section{Proof of Theorem~\ref{ThmAngle} for two clusters}

Due to Theorem \ref{ThmStable}, there exists a $t_0\geq 1$ such that for every $t\geq t_0$, every row/neuron of $X^{(t)}$ is a vector composed of at most 3 clusters. To improve the readability and ease of understanding of the proof, we first prove the theorem neglecting the presence of three-cluster rows. This is because the main ideas are already present in this simpler case, and three-cluster rows only introduce some unenlightening technicalities to the proof. We address these technicalities in Appendix \ref{3ClusterProof}, where we prove that three-cluster rows do not invalidate the result.

To prove the first point of the theorem, notice that, due to the fact that the components of each row of $X^{(t)}$ are ordered, each cluster is composed of consecutive elements. Hence, the first and the last element of each row are always in two different clusters: either the first element is zero and the last is strictly positive, or vice versa. Hence, the product of these two elements is always zero, and therefore $\langle \bx_1^{(t)},\bx_n^{(t)}\rangle = 0$.

To prove the second part of the theorem, consider any row of $X^{(t)}$ that is composed of two clusters, and recall that one of the two clusters must be at zero, and the other one at a positive coordinate. We will refer to those clusters as the \emph{zero cluster} and the \emph{positive cluster} respectively. After $T$ additional layers, the selected row undergoes any possible sequence of $T$ positive and/or negative transformations, generating a total of $2^T$ rows in the matrix $X^{(t+T)}$. We want to study the contribution of these $2^T$ rows to the inner product between any two columns of $X^{(t+T)}$.

Let $i,j \in \{2,\dots,n-1\}$ be the two selected columns. First of all, notice that the $2^T$ rows generated by a given row at layer $t$, will contribute to the inner product between $i$ and $j$ only if elements $i$ and $j$ of the given row are in the same cluster at layer $t$; otherwise, one of the two elements will necessarily be zero. Furthermore, of the corresponding $2^T$ rows at layer $t+T$, those that contribute to the inner product are only the ones in which the cluster containing $i$ and $j$ is the positive one. Hence, the given neuron at layer $t$ has to undergo a sequence of positive/negative transformations in such a way that after $T$ layers, the cluster containing $i$ and $j$ is the positive one.

\begin{figure}
\includegraphics[width=0.6\textwidth]{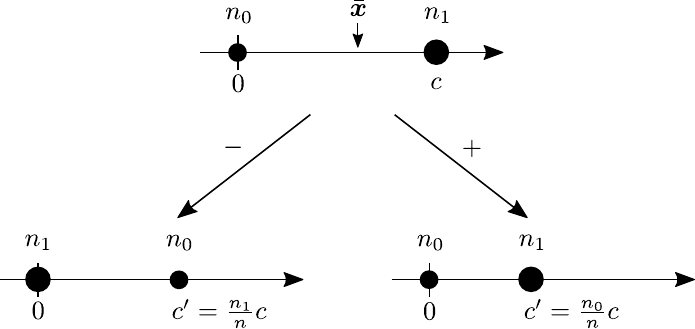}
\centering
\caption{Example of evolution of stable configurations with two clusters. The number of points in the two clusters are denoted by $n_0$ and $n_1$.}
\label{fig:two-cluster}
\end{figure}

Keeping all of this in mind, the next step is to study how a two-cluster vector changes after a given sequence of $T$ positive/negative transformations. Consider any vector of $n$ elements composed of two clusters: one cluster is at coordinate 0 (the zero cluster) and it is made of $n_0$ elements, while the second cluster (the positive cluster) is at coordinate $c >0$ and it is made of $n-n_0$ elements. From now on, we say that such a vector has \emph{composition} $(n_0, n-n_0)$. After one positive transformation, the zero cluster remains unchanged, while the coordinate of the positive cluster becomes
\begin{align}
    c' = c - \frac{n-n_0}{n}c = \frac{n_0}{n}c.
\end{align}
After a negative transformation, instead, the zero and positive cluster switch roles: the positive cluster goes at coordinate 0 and becomes the new zero cluster, while the old zero cluster becomes the new positive cluster. Thus, the new vector has composition $(n-n_0,n_0)$, and the new coordinate of the positive cluster is equal to
\begin{align}
    c' = \frac{n-n_0}{n}c.
\end{align}
See Figure~\ref{fig:two-cluster} for a visual representation of this process.
Thus, one can see that what really determines the new coordinate of the positive cluster after one transformation is the number of points in the zero cluster \emph{after} the transformation: in the first case, the number of points is $n_0$, while in the second case it is $n-n_0$. After $T$ transformations, in which for $k$ times the zero cluster contains $n_0$ points, and for $T-k$ times it contains $n-n_0$ points, the final coordinate of the positive cluster is
\begin{align}
\label{TTransfCoord}
    c' = \bigg(\frac{n_0}{n}\bigg)^k \left(\frac{n-n_0}{n}\right)^{T-k} c.
\end{align}
Hence, for a row $\rx_{\ell}^{(t)}$ at layer $t$ in which elements $i$ and $j$ are in the same cluster with $n_{\ell}$ elements, all the neurons generated from it after a sequence of $T$ transformations in which their cluster ends in a positive coordinate $c$, contributes to the inner product, with a value equal to $c^2$. Following \eqref{TTransfCoord}, the total contribution is
%\IN{explanation why it ends with T-1. Also, no equality fro 22-> 23, right?}
\begin{align}
    C_{\ell} &= \sum_{k=0}^{T-1}\binom{T-1}{k}\left(\frac{n-n_{\ell}}{n}\right)^{2(T-k)}\bigg(\frac{n_{\ell}}{n}\bigg)^{2k} c_{\ell}^2 \\
        &= c_{\ell}^2 \left(\frac{n-n_{\ell}}{n}\right)^2 \left\{\bigg(\frac{n-n_{\ell}}{n}\bigg)^2 + \bigg(\frac{n_{\ell}}{n}\bigg)^2\right\}^{T-1} \label{TwoClusterContrib}
\end{align}
where $c_{\ell}$ is the coordinate of the positive cluster of $\rx_{\ell}^{(t)}$. Note that the leading term $(\frac{n-n_{\ell}}{n})^2$ and the $T-1$ exponent appear because, after any sequence of $T-1$ transformations, the $T$-th and last one is fixed, since the cluster containing elements $i$ and $j$ must end in a positive coordinate.
As one can see, all contributions decrease exponentially with $T$, and the exponent depends on the composition of the two clusters. The compositions that dominate are the most unbalanced ones: those with one point in one cluster and $n-1$ points in the other. Thus, we can limit our attention to the leaves belonging to the set
\begin{multline}
\label{LijClusters}
    L^{(t)}_{i,j} = \{\ell:  \rx_{\ell}^{(t)} \text{ has two clusters of composition } (1,n-1) \text{ or } (n-1,1)\\\text{with elements }i,j \text{ in the same cluster}\}
\end{multline}
since, asymptotically as $T$ goes to infinity, these are the only ones that contribute to the inner product. Note that for any pair $i,j$, such unbalanced configurations always exist, since there is always a path of positive/negative transformations that leads to them. Hence, the set $L^{(t)}_{i,j}$ is not empty. Furthermore, if a cluster contains $n-1$ elements, then these elements must be either $\{2,...n\}$ or $\{1,\dots,n-1\}$. Hence, all elements in $\{2,\dots,n-2\}$ must belong to the same cluster. Thus, the set defined in \eqref{LijClusters} contains the same neurons for any pair $i,j\in\{2,\dots,n-1\}$, and we can define a single set independent of $i,j$:
\begin{align}
    L^{(t)} = \{\ell: \rx_{\ell}^{(t)} \text{ has two clusters of composition } (1,n-1) \text{ or } (n-1,1)\}
\end{align}
so that $L^{(t)}_{i,j} = L^{(t)}$ for every $i,j\in\{2,\dots,n\}$.
The total contribution of the clusters that belong to $L^{(t)}$ amounts to %\IN{mentioning that such unbalanced ones exist for all pairs i,j}
\begin{align}
\label{TotalContrib}
    \sum_{\ell\in L^{(t)}} C_{\ell} = \frac{1}{n^2}\left\{\bigg(\frac{1}{n}\bigg)^2 + \bigg(\frac{n-1}{n}\bigg)^2\right\}^{T-1} \sum_{\ell\in L^{(t)}} c_{\ell}^2
\end{align}
which is independent of the actual choice of $i$ and $j$.

Regarding the norm of a single element $i$, the same reasoning applies, with the difference that the rows that contribute to it are all those in which $i$ is in the positive cluster. Again, the dominant configurations are the most unbalanced ones, and since $i$ and $j$ are not extreme points, the dominant configurations that contribute to the norms $\| \bx_i^{(t+T)}\|$ and $\| \bx_j^{(t+T)}\|$ are exactly the same that contribute to $\langle \bx_i^{(t+T)},\bx_j^{(t+T)}\rangle$, i.e., those in $L^{(t)}$. Thus, we have
% \begin{align}
%     \| X_{t+T,i}\| \simeq \| X_{t+T,j}\| \simeq \sqrt{\frac{1}{n}\left\{\bigg(\frac{1}{n}\bigg)^2 + \bigg(\frac{n-1}{n}\bigg)^2\right\}^{T-1} \sum_{\ell\in L^{(t)}_{i,j}} c_{\ell}^2}
% \end{align}
%\IN{writing explicit formula}
\begin{multline}
\label{AngleLimit}
    \lim_{T\to\infty} \frac{\langle \bx_i^{(t+T)},\bx_j^{(t+T)}\rangle}{\| \bx_i^{(t+T)}\| \| \bx_j^{(t+T)}\|} \\
        = \lim_{T\to\infty} \frac{\frac{1}{n^2}\left\{(\frac{1}{n})^2 + (\frac{n-1}{n})^2\right\}^{T-1} \sum_{\ell\in L^{(t)}} c_{\ell}^2 + o\left(\left((\frac{1}{n})^2 + (\frac{n-1}{n})^2\right)^T\right)}{\frac{1}{n^2}\left\{(\frac{1}{n})^2 + (\frac{n-1}{n})^2\right\}^{T-1} \sum_{\ell\in L^{(t)}} c_{\ell}^2 +o\left(\left((\frac{1}{n})^2 + (\frac{n-1}{n})^2\right)^T\right)} = 1.
\end{multline}
where $\frac{o\left(\left((\frac{1}{n})^2 + (\frac{n-1}{n})^2\right)^T\right)}{\left((\frac{1}{n})^2 + (\frac{n-1}{n})^2\right)^T} \to 0$ as $T\to\infty$.

As shown in Appendix A, the contribution of three-cluster configurations does not change the validity of \eqref{AngleLimit}, so the second part of the theorem is proved.

To prove the third part, recall that, from the previous discussion, we showed that, asymptotically as $T\to\infty$, the norm of any vector $\bx_i^{(t+T)}$ for $i\in\{2,\dots,n-1\}$ is equal to, following \eqref{TotalContrib},
\begin{align}
\label{NormXi}
\|\bx_i^{(t+T)}\|^2 \simeq \frac{1}{n^2}\left\{\bigg(\frac{1}{n}\bigg)^2 + \bigg(\frac{n-1}{n}\bigg)^2\right\}^{T-1} \sum_{\ell\in L^{(t)}} c_{\ell}^2.
\end{align}
Next, notice that the vectors belonging to $L^{(t)}$ can have two different compositions, namely $(1,n-1)$ and $(n-1,1)$. While the rows with these two compositions have the same contribution to the norm of a vector with index between 2 and $n-1$, this is not true anymore for the extreme vectors $\bx_1^{(t+T)}$ and $\bx_n^{(t+T)}$. In fact, for a row at layer $t$ with composition $(1,n-1)$, out of the $2^T$ rows generated by it at layer $t+T$, those that contribute to  $\|\bx_1^{(t+T)}\|$ and those that contribute to $\|\bx_i^{(t+T)}\|$ for $i\in\{2,\dots,n-1\}$ are complementary sets. If instead the starting configuration is $(n-1,1)$, then the rows that contribute to the two norms are exactly the same, since points 1 and $i$ belong to the same cluster. A similar reasoning applies also to $\|\bx_n^{(t+T)}\|$.

Let us call $L^{(t)}_1$ the set of rows at layer $t$ with composition $(1,n-1)$, and $L^{(t)}_2$ those with composition $(n-1,1)$. Then, $L^{(t)} = L^{(t)}_1 \cup L^{(t)}_2$. Following \eqref{TwoClusterContrib}, the total contribution to $\|\bx_1^{(t+T)}\|$ of the rows in $L^{(t)}$ is
\begin{align}
\label{NormX1}
\|\bx_1^{(t+T)}\|^2 \simeq \left\{\bigg(\frac{1}{n}\bigg)^2 + \bigg(\frac{n-1}{n}\bigg)^2\right\}^{T-1} \left(\left(\frac{n-1}{n}\right)^2\sum_{\ell\in L^{(t)}_1} c_{\ell}^2 + \left(\frac{1}{n}\right)^2 \sum_{\ell\in L^{(t)}_2} c_{\ell}^2\right).
\end{align}
Similarly, one has
\begin{align}
\label{NormXn}
    \|\bx_n^{(t+T)}\|^2 \simeq \left\{\bigg(\frac{1}{n}\bigg)^2 + \bigg(\frac{n-1}{n}\bigg)^2\right\}^{T-1} \left(\left(\frac{1}{n}\right)^2\sum_{\ell\in L^{(t)}_1} c_{\ell}^2 + \left(\frac{n-1}{n}\right)^2 \sum_{\ell\in L^{(t)}_2} c_{\ell}^2\right).
\end{align}
Hence, we can write
\begin{align}
    \frac{\|\bx_i^{(t+T)}\|^2}{\|\bx_1^{(t+T)}\|^2} &\simeq \frac{\left(\frac{1}{n}\right)^2 \sum_{\ell\in L^{(t)}} c_{\ell}^2}{\left(\frac{n-1}{n}\right)^2\sum_{\ell\in L^{(t)}_1} c_{\ell}^2 + \left(\frac{1}{n}\right)^2 \sum_{\ell\in L^{(t)}_2} c_{\ell}^2} \\
    &= \frac{1+ \frac{\sum_{\ell\in L^{(t)}_2} c_{\ell}^2}{\sum_{\ell\in L^{(t)}_1} c_{\ell}^2}}{(n-1)^2 + \frac{\sum_{\ell\in L^{(t)}_2} c_{\ell}^2}{\sum_{\ell\in L^{(t)}_1} c_{\ell}^2}}
\end{align}
and, similarly,
\begin{align}
    \frac{\|\bx_i^{(t+T)}\|^2}{\|\bx_n^{(t+T)}\|^2} \simeq \frac{1+ \frac{\sum_{\ell\in L^{(t)}_1} c_{\ell}^2}{\sum_{\ell\in L^{(t)}_2} c_{\ell}^2}}{(n-1)^2 + \frac{\sum_{\ell\in L^{(t)}_1} c_{\ell}^2}{\sum_{\ell\in L^{(t)}_2} c_{\ell}^2}}.
\end{align}
Therefore, if $\frac{\sum_{\ell\in L^{(t)}_2} c_{\ell}^2}{\sum_{\ell\in L^{(t)}_1} c_{\ell}^2} \leq 1$, then
\begin{align}
\label{NormRatioBound1}
    \frac{\|\bx_i^{(t+T)}\|^2}{\|\bx_1^{(t+T)}\|^2} \lesssim \frac{2}{(n-1)^2}.
\end{align}
If instead $\frac{\sum_{\ell\in L^{(t)}_2} c_{\ell}^2}{\sum_{\ell\in L^{(t)}_1} c_{\ell}^2} > 1$, then
\begin{align}
\label{NormRatioBound2}
    \frac{\|\bx_i^{(t+T)}\|^2}{\|\bx_n^{(t+T)}\|^2} \lesssim \frac{2}{(n-1)^2}.
\end{align}
As shown in Appendix A, the contribution of three-cluster configurations makes the bounds in \eqref{NormRatioBound1} and \eqref{NormRatioBound2} larger (as in the statement of the theorem), but does not change the general validity of the result.

The proof of the fourth part of the theorem is similar to the third. In fact, we can write
\begin{align}
    \langle \bx_i^{(t+T)}, \bx_1^{(t+T)}\rangle \simeq \left(\frac{1}{n}\right)^2\left\{\bigg(\frac{1}{n}\bigg)^2 + \bigg(\frac{n-1}{n}\bigg)^2\right\}^{T-1} \sum_{\ell\in L^{(t)}_2} c_{\ell}^2
\end{align}
and
\begin{align}
    \langle \bx_i^{(t+T)}, \bx_n^{(t+T)}\rangle \simeq \left(\frac{1}{n}\right)^2\left\{\bigg(\frac{1}{n}\bigg)^2 + \bigg(\frac{n-1}{n}\bigg)^2\right\}^{T-1} \sum_{\ell\in L^{(t)}_1} c_{\ell}^2
\end{align}
Hence, using also \eqref{NormX1}, \eqref{NormXn} and \eqref{NormXi}, one has
\begin{align}
    \phase{\bx_i^{(t+T)}}{\bx_1^{(t+T)}} &= \arccos\frac{\langle \bx_i^{(t+T)},\bx_1^{(t+T)}\rangle}{\| \bx_i^{(t+T)}\| \| \bx_1^{(t+T)}\|} \\
    &\simeq \arccos\frac{\sum_{\ell\in L^{(t)}_2} c_{\ell}^2}{\sqrt{\sum_{\ell\in L^{(t)}_1} c_{\ell}^2 + \sum_{\ell\in L^{(t)}_2} c_{\ell}^2}\sqrt{(n-1)^2 \sum_{\ell\in L^{(t)}_1} c_{\ell}^2 + \sum_{\ell\in L^{(t)}_2} c_{\ell}^2}} \\
    &= \arccos\frac{1}{\sqrt{\frac{\sum_{\ell\in L^{(t)}_1} c_{\ell}^2}{\sum_{\ell\in L^{(t)}_2} c_{\ell}^2}+1}\sqrt{(n-1)^2\frac{\sum_{\ell\in L^{(t)}_1} c_{\ell}^2}{\sum_{\ell\in L^{(t)}_2} c_{\ell}^2} + 1}}
\end{align}
and similarly,
\begin{align}
    \phase{\bx_i^{(t+T)}}{\bx_n^{(t+T)}} \simeq \arccos\frac{1}{\sqrt{\frac{\sum_{\ell\in L^{(t)}_2} c_{\ell}^2}{\sum_{\ell\in L^{(t)}_1} c_{\ell}^2}+1}\sqrt{(n-1)^2\frac{\sum_{\ell\in L^{(t)}_2} c_{\ell}^2}{\sum_{\ell\in L^{(t)}_1} c_{\ell}^2} + 1}}.
\end{align}
Therefore, if $\frac{\sum_{\ell\in L^{(t)}_2} c_{\ell}^2}{\sum_{\ell\in L^{(t)}_1} c_{\ell}^2} \leq 1$, then
\begin{align}
    \phase{\bx_i^{(t+T)}}{\bx_1^{(t+T)}} \gtrsim \arccos\frac{1}{\sqrt{2}(n-1)} \geq \frac{\pi}{2} - \frac{\pi}{2\sqrt{2}(n-1)}.
\end{align}
Otherwise, 
\begin{align}
    \phase{\bx_i^{(t+T)}}{\bx_n^{(t+T)}} \gtrsim \frac{\pi}{2} - \frac{\pi}{2\sqrt{2}(n-1)}.
\end{align}
Once again, three-cluster configurations slightly change the value of the bound, as shown in Appendix A.

\section{Contribution of three-cluster configurations to Theorem \ref{ThmAngle}}
\label{3ClusterProof}

In Theorem \ref{ThmAngle}, rows at layer $t$ which are composed of three clusters must also be taken into account. To complete the proof of the second point of the theorem, we now show that their total contribution to the numerator and the denominator in \eqref{AngleDef} are asymptotically equal, and therefore \eqref{AngleLimit} still holds in the general case.

First of all, for the same reasoning as in the two-cluster case, a three-cluster row at layer $t$ will have a non-negligible contribution at layer $t+T$ only if at some point during the process between layer $t$ and layer $t+T$, it generates two-cluster rows with composition $(1,n-1)$ or $(n-1,1)$; otherwise, its contribution will be negligible compared to the contribution of the two-cluster configurations belonging to $L^{(t)}$. Such three-cluster rows must have composition $(1,1,n-2)$ or $(n-2,1,1)$ --- the cluster in the middle must contain only one element by construction. Furthermore, once again by construction, a three-cluster row at layer $t$ generates $2^T$ rows at layer $T$, out of which exactly one is composed of three-clusters row, and the other $2^T - 1$ are two-cluster ones.

The contribution of the single three-cluster row is negligible, and this can be proved even by using very loose bounds. In fact, for any three-cluster configuration, let $c$ be the coordinate of the largest cluster. Then, after one layer, both the new coordinates of the middle and the largest clusters can be upper bounded by $\frac{n-1}{n}c$. Thus, after $T$ layers, the final coordinates are upper bounded by $\left(\frac{n-1}{n}\right)^T c$. Hence, its contribution to the inner products or to the norms will be at most $\left(\frac{n-1}{n}\right)^{2T} c^2$, which has a negligible exponent compared to \eqref{TwoClusterContrib}.
%\IN{need better name than rows?}

\begin{figure}
\includegraphics[width=0.6\textwidth]{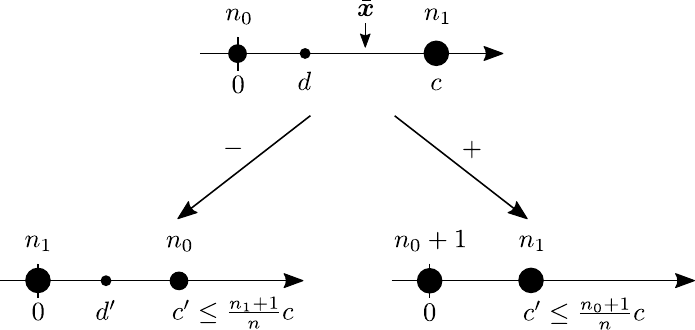}
\centering
\caption{Example of evolution of stable configurations with three clusters. Notice that, at every step, a three-cluster neuron always generates one three-cluster neuron and one two-cluster neuron.}
\label{fig:threeclusters}
\end{figure}

Regarding the contribution of the other rows, first notice that, if the elements $i$ and $j$ are both in the largest cluster at layer $t$, they will remain paired (i.e., in the same cluster) in all the rows generated at layer $t+T$. Thus, as before, the contribution of the rows will be exactly the same in the inner product at the numerator of \eqref{AngleDef}, and in the product of the norms in the denominator. The remaining case is that in which one element is in the middle cluster, and the other is in the largest cluster. Notice that, at each step, a three-cluster row generates two rows: one has a three-cluster configuration, while the other one has two clusters, with composition $(n-1,1)$ or $(2,n-2)$ (see Figure \ref{fig:threeclusters} for a visual example). Thus, at layer $t+T$, the two-cluster rows have composition $(n-1,1)$, $(1,n-1)$, $(2,n-2)$ or $(n-2,2)$. The rows with compositions $(n-1,1)$ or $(1,n-1)$ give, once again, the same contribution to the numerator and the denominator of \eqref{AngleDef}. Instead, those with composition $(2,n-2)$ or $(n-2,2)$ give different contributions: they give zero contribution to the inner product, since the two elements under consideration belong to different clusters, but they contribute to each of the norms. However, asymptotically their total contribution is negligible. In fact, the contribution to both $\| \bx_i^{(t+T)}\|^2$ and $\| \bx_j^{(t+T)}\|^2$ can be upper bounded, thanks to \eqref{TwoClusterContrib}, by
\begin{align}
    c^2 \left(\frac{n-2}{n}\right)^2&\Bigg\{\left(\frac{n-1}{n}\right)^2 \left[\left(\frac{2}{n}\right)^2 + \left(\frac{n-2}{n}\right)^2\right]^{T-2} + \left(\frac{n-1}{n}\right)^4 \left[\left(\frac{2}{n}\right)^2 + \left(\frac{n-2}{n}\right)^2\right]^{T-3} \notag\\
        &\hspace{20em}+\cdots + \left(\frac{n-1}{n}\right)^{2(T-1)}\Bigg\}\\
        &\leq c^2 \left(\frac{n-2}{n}\right)^2 \left(\frac{n^2-2n+1}{2n-7}\right)\left(\frac{n-1}{n}\right)^{2(T-1)} \label{NegligibleContrib}
\end{align}
which has a negligible exponent with respect to \eqref{TwoClusterContrib}. Hence, we proved that three-cluster configurations do not affect the validity of \eqref{AngleLimit}, and the second part of the theorem is proved.

For the third part of the theorem, the three-cluster configurations that contribute to the norms are again only those of the form $(1,1,n-2)$ or $(n-2,1,1)$, since those are the only non-negligible ones. First, let us consider a three-cluster row at layer $t$ of composition $(1,1,n-2)$ and an element $i \in \{3,n-2\}$. We want to study the cumulative contribution at layer $t+T$ of all the rows generated by the selected one at layer $t$. Notice that, at layer $t+T-1$, one generated row is of composition $(1,1,n-2)$, and the others have composition $(1,n-1)$ or $(2,n-2)$. The contribution of the single three-cluster row is negligible, as we already discussed above. For a row of composition $(1,n-1)$, let $c(T)$ be the value of the coordinate of the positive cluster at layer $t+T-1$ (which depends exponentially on $T$). Of its two children at layer $t+T$, one contributes to the norm of $\bx_i^{(t+T)}$ and $\bx_n^{(t+T)}$ an amount equal to $\frac{1}{n^2}c^2(T)$, and the other one to the norm of $\bx_1^{(t+T)}$ an amount equal to $\left(\frac{n-1}{n}\right)^2 c^2(T)$. Similarly, for a row of composition $(2,n-2)$, one contributes to the norm of $\bx_i^{(t+T)}$ and $\bx_n^{(t+T)}$ an amount equal to $\frac{4}{n^2}c^2(T)$, and the other one to the norm of $\bx_1^{(t+T)}$ an amount equal to $\left(\frac{n-2}{n}\right)^2 c^2(T)$. A similar reasoning can be applied to three-cluster neurons at layer $t$ with composition $(n-2,1,1)$. Putting everything together, define $A(T)$ to be the sum of the squares of the positive coordinates of all rows of composition $(1,n-1)$ at layer $t+T-1$, let $B(T)$ be the same for rows of composition $(n-1,1)$, $C(T)$ for rows of composition $(2,n-2)$, and $D(T)$ for rows of composition $(n-2,2)$. Then, we have:
\begin{align}
    \|\bx_i^{(t+T)}\|^2 &\simeq \frac{1}{n^2}A(T) + \frac{1}{n^2}B(T) + \frac{4}{n^2}C(T) + \frac{4}{n^2}D(T) \\
    \|\bx_1^{(t+T)}\|^2 &\simeq \left(\frac{n-1}{n}\right)^2 A(T) + \frac{1}{n^2}B(T) + \left(\frac{n-2}{n}\right)^2 C(T) + \frac{4}{n^2}D(T) \\
    \|\bx_n^{(t+T)}\|^2 &\simeq \frac{1}{n^2}A(T) + \left(\frac{n-1}{n}\right)^2 B(T) + \frac{4}{n^2}C(T) + \left(\frac{n-2}{n}\right)^2 D(T)
\end{align}
from which it follows that
\begin{align}
    \frac{\|\bx_i^{(t+T)}\|^2}{\|\bx_1^{(t+T)}\|^2} &\simeq \frac{A(T)+B(T)+4C(T)+4D(T)}{(n-1)^2 A(T)+B(T)+(n-2)^2 C(T)+4D(T)} \\
        &\leq \frac{4(A(T)+C(T)) + 4(B(T)+D(T))}{(n-2)^2(A(T)+C(T)) + (B(T)+D(T))} \\
        &= \frac{4 + 4\frac{B(T)+D(T)}{A(T)+C(T)}}{(n-2)^2 + \frac{B(T)+D(T)}{A(T)+C(T)}}
\end{align}
and
\begin{align}
    \frac{\|\bx_i^{(t+T)}\|^2}{\|\bx_n^{(t+T)}\|^2} &\simeq \frac{A(T)+B(T)+4C(T)+4D(T)}{A(T)+(n-1)^2B(T)+4C(T)+(n-2)^2D(T)} \\
        &\leq \frac{4(A(T)+C(T)) + 4(B(T)+D(T))}{(n-2)^2(B(T)+D(T)) + (A(T)+C(T))} \\
        &= \frac{4 + 4\frac{A(T)+C(T)}{B(T)+D(T)}}{(n-2)^2 + \frac{A(T)+C(T)}{B(T)+D(T)}}
\end{align}
since $A(T),B(T),C(T)$ and $D(T)$ are all positive for every $T\geq 1$. Furthermore, $\frac{A(T)+C(T)}{B(T)+D(T)}$ converges to a constant as $T\to\infty$. Hence, for every $T$ larger than a certain value, either $\frac{A(T)+C(T)}{B(T)+D(T)} \leq 1$ or $\frac{B(T)+D(T)}{A(T)+C(T)} \leq 1$. From this, it follows that either 
\begin{align}
    \lim_{t\to\infty} \frac{\|\bx_i^{(t)}\|^2}{\|\bx_1^{(t)}\|^2} \leq \frac{8}{(n-2)^2}
\end{align}
or
\begin{align}
    \lim_{t\to\infty} \frac{\|\bx_i^{(t)}\|^2}{\|\bx_n^{(t)}\|^2} \leq \frac{8}{(n-2)^2}
\end{align}
which proves the theorem, for the case of $i\in\{3,\dots,n-2\}$. To complete the proof of part three, the case of points $i=2$ and $i=n-1$ must be considered. We analyze the case $i=2$, since the other case follows in the same way by symmetry. Let $\ell$ be any three-cluster row at layer $t$ with composition $(1,1,n-2)$. During the next $T$ layers, point $i=2$ will get paired with point 1 only in $(2,n-2)$ cluster configurations, whose contribution compared to $(1,n-1)$ ones is negligible, per \eqref{NegligibleContrib}. The only other two-cluster configurations generated during the process are $(1,n-1)$ ones, in which point 2 is always paired with point $n$. Thus, for those clusters, due to \eqref{TwoClusterContrib}, the final contribution at layer $t+T$ will be of the form $\frac{1}{n^2} e_{\ell}(T)$ for the norms $\|\bx_2^{(t+T)}\|^2$ and $\|\bx_n^{(t+T)}\|^2$, and of the form $\left(\frac{n-1}{n}\right)^2 e_{\ell}(T)$ for $\|\bx_1^{(t+T)}\|^2$, where $e_{\ell}(T)$ is an exponentially-decaying function of $T$, whose exponent can be at most $\left\{\left(\frac{1}{n}\right)^2+\left(\frac{n-1}{n}\right)^2\right\}^{T-1}$. Denote by $C(T)$ the cumulative exponent for all $(1,1,n-2)$ rows at layer $t$, i.e., $C(T) = \sum_{\ell} e_{\ell}(T)$, where the sum is over all three-cluster rows at layer $t$ with composition $(1,1,n-2)$. On the contrary, for three-cluster rows with composition $(n-2,1,1)$, point 2 is always paired with point $1$. Hence, if $\ell$ is such a row, the final contribution at layer $t+T$ will be of the form $\frac{1}{n^2} e_{\ell}(T)$ for the norms $\|\bx_1^{(t+T)}\|^2$ and $\|\bx_2^{(t+T)}\|^2$, and of the form $\left(\frac{n-1}{n}\right)^2 e_{\ell}(T)$ for $\|\bx_n^{(t+T)}\|^2$. Denote by $D(T)$ the cumulative exponent for all $(1,1,n-2)$ rows at layer $t$. Furthermore, denote by $A(T)$ and $B(T)$ the cumulative exponent for two-cluster rows with composition $(1,n-1)$ and $(n-1,1)$ respectively, whose analysis was carried out in the proof of the two-cluster case in the main section. Following the discussion above, we get the following asymptotic formulas for large $T$,
\begin{align}
    \|\bx_1^{(t+T)}\|^2 &\simeq \left(\frac{n-1}{n}\right)^2 A(T) + \frac{1}{n^2}B(T) + \left(\frac{n-1}{n}\right)^2 C(T) + \frac{1}{n^2}D(T) \\
    \|\bx_2^{(t+T)}\|^2 &\simeq \frac{1}{n^2}A(T) + \frac{1}{n^2}B(T) + \frac{1}{n^2}C(T) + \frac{1}{n^2}D(T) \\
    \|\bx_n^{(t+T)}\|^2 &\simeq \frac{1}{n^2}A(T) + \left(\frac{n-1}{n}\right)^2 B(T) + \frac{1}{n^2}C(T) + \left(\frac{n-1}{n}\right)^2 D(T)
\end{align}
from which it follows that, for large $T$,
\begin{align}
\frac{\|\bx_2^{(t+T)}\|^2}{\|\bx_1^{(t+T)}\|^2} &\simeq \frac{1 + \frac{B(T)+D(T)}{A(T)+C(T)}}{(n-1)^2 + \frac{B(T)+D(T)}{A(T)+C(T)}}\\
\frac{\|\bx_2^{(t+T)}\|^2}{\|\bx_n^{(t+T)}\|^2} &\simeq \frac{1 + \frac{A(T)+C(T)}{B(T)+D(T)}}{(n-1)^2 + \frac{A(T)+C(T)}{B(T)+D(T)}}.
\end{align}
As before, if $\lim_{T\to\infty}\frac{A(T)+C(T)}{B(T)+D(T)} \geq 1$, then
\begin{align}
    \lim_{t\to\infty}\frac{\|\bx_2^{(t)}\|^2}{\|\bx_1^{(t)}\|^2} \leq \frac{2}{(n-1)^2},
\end{align}
otherwise,
\begin{align}
    \lim_{t\to\infty}\frac{\|\bx_2^{(t)}\|^2}{\|\bx_n^{(t)}\|^2} \leq \frac{2}{(n-1)^2}.
\end{align}

The proof for the fourth part follows similarly. Using the same notation, one has, for $i\in\{3,n-2\}$,
\begin{align}
    \langle \bx_i^{(t+T)}, \bx_1^{(t+T)}\rangle \simeq \frac{1}{n^2} B(T) + \frac{4}{n^2}D(T)
\end{align}
and
\begin{align}
    \langle \bx_i^{(t+T)}, \bx_n^{(t+T)}\rangle \simeq \frac{1}{n^2} A(T) + \frac{4}{n^2}C(T)
\end{align}
from which it follows that
\begin{align}
    \frac{\langle \bx_i^{(t+T)},\bx_1^{(t+T)}\rangle}{\| \bx_i^{(t+T)}\| \| \bx_1^{(t+T)}\|}&= \\
    &\hspace{-8em}\simeq \frac{B(T)+4D(T)}{\sqrt{A(T)+B(T)+4C(T)+4D(T)}\sqrt{(n-1)^2 A(T) + B(T) + (n-2)^2 C(T) + 4D(T)}} \\
    &\hspace{-8em}\leq \frac{4}{\sqrt{\frac{A(T)+C(T)}{B(T)+D(T)}+1}\sqrt{(n-2)^2\frac{A(T)+C(T)}{B(T)+D(T)}+1}}
\end{align}
and
\begin{align}
    \frac{\langle \bx_i^{(t+T)},\bx_n^{(t+T)}\rangle}{\| \bx_i^{(t+T)}\| \| \bx_n^{(t+T)}\|} \lesssim \frac{4}{\sqrt{\frac{B(T)+D(T)}{A(T)+C(T)}+1}\sqrt{(n-2)^2\frac{B(T)+D(T)}{A(T)+C(T)}+1}}
\end{align}
As before, for every $T$ larger than a certain value, either $\frac{A(T)+C(T)}{B(T)+D(T)} \leq 1$ or $\frac{B(T)+D(T)}{A(T)+C(T)} \leq 1$, from which we can conclude that either
\begin{align}
    \phase{\bx_i^{(t)}}{\bx_1^{(t)}} \gtrsim \frac{\pi}{2} - \frac{\sqrt{2}\pi}{n-2}
\end{align}
or
\begin{align}
    \phase{\bx_i^{(t)}}{\bx_n^{(t)}} \gtrsim \frac{\pi}{2} - \frac{\sqrt{2}\pi}{n-2}.
\end{align}
For the case $i=2$ (the case $i=n-1$ follows again by symmetry), let again $A(T)$, $B(T)$, $C(T)$ and $D(T)$ be the cumulative exponents for clusters with composition $(1,n-1)$, $(n-1,1)$, $(1,1,n-2)$, $(n-2,1,1)$ respectively. Notice that clusters with composition $(1,1,n-2)$ asymptotically contribute only to $\langle X_i^{(t+T)}, X_n^{(t+T)}\rangle$, since point 1 and 2 get paired only when clusters of composition $(2,n-2)$ are formed in the process from layer $t$ to $t+T$, which are asymptotically negligible per \eqref{NegligibleContrib}. The opposite is true for clusters of composition $(n-2,1,1)$. Hence, asymptotically for large $T$, we have the formulae
\begin{align}
    \langle \bx_2^{(t+T)}, \bx_1^{(t+T)}\rangle &\simeq \frac{1}{n^2} B(T) + \frac{1}{n^2}D(T) \\
    \langle \bx_2^{(t+T)}, \bx_n^{(t+T)}\rangle &\simeq \frac{1}{n^2} A(T) + \frac{1}{n^2}C(T)
\end{align}
from which it follows that
\begin{align}
    \frac{\langle \bx_2^{(t+T)},\bx_1^{(t+T)}\rangle}{\| \bx_2^{(t+T)}\| \| \bx_1^{(t+T)}\|}&\simeq \frac{1}{\sqrt{1+(n-1)^2\frac{A(T)+C(T)}{B(T)+D(T)}}\sqrt{1 + \frac{A(T)+C(T)}{B(T)+D(T)}}} \\
    \frac{\langle \bx_2^{(t+T)},\bx_n^{(t+T)}\rangle}{\| \bx_2^{(t+T)}\| \| \bx_1^{(t+T)}\|}&\simeq \frac{1}{\sqrt{1+(n-1)^2\frac{B(T)+D(T)}{A(T)+C(T)}}\sqrt{1 + \frac{B(T)+D(T)}{A(T)+C(T)}}}.
\end{align}
Once again, if $\lim_{T\to\infty}\frac{A(T)+C(T)}{B(T)+D(T)} \geq 1$, then
\begin{align}
    \lim_{t\to\infty}\frac{\langle \bx_2^{(t)},\bx_1^{(t)}\rangle}{\| \bx_2^{(t)}\| \| \bx_1^{(t)}\|} \leq \frac{1}{\sqrt{2}(n-1)}.
\end{align}
Otherwise,
\begin{align}
    \lim_{t\to\infty}\frac{\langle \bx_2^{(t)},\bx_n^{(t)}\rangle}{\| \bx_2^{(t)}\| \| \bx_n^{(t)}\|} \leq \frac{1}{\sqrt{2}(n-1)}.
\end{align}
This concludes the proof of the theorem for the general case.

\section{Proof of Theorem~\ref{thm:invariant} and Theorem~\ref{thm:rand_stb}}
For both theorems, it is sufficient to calculate the expectation for a single row $\rw$ of $W$. This holds since the quantities of interest may be separated as a sum, where each summand corresponds to a unique row, and since all the rows are i.i.d\, each row has an equal contribution. For example, let us start by showing that norms of vectors may be scaled  uniformly to our desire, in expectation, when choosing a suitable $\sigma$.

\begin{align}\label{eq:norm}
   \EE{} \norm{\R \lrp{W  \bx } }^2  &=  \EE{}  \R \lrp{W  \bx } \cdot \R \lrp{W  \bx } \\ 
   &= \EE{} \sum_{i=1}^d \R \lrp{\rw_i    {\bx}  }  \R \lrp{\rw_i    {\bx} } \\
    &=  \sum_{i=1}^d \EE{} \R \lrp{\rw    {\bx}  }  \R \lrp{\rw    {\bx} } \\
    &=   \frac{d\sigma^2}{2}  \norm{ \bx }^2 
\end{align}
The second equality follows from the definition of the scalar product, and the third equality is by the linearity of expectation. The fourth equality is less trivial and follows by noticing that $N \coloneqq \rw  \bx \sim \mathcal{N}\lrp{0,\sigma^2\norm{\bx}^2}$. Since $\mathrm{ReLU}(x) = 0$ for negative $x$, we get that $\EE{} \R(N)^2 = \frac{1}{2} \EE{} N^2 = \frac{\sigma^2 \norm{\bx}^2}{2}$.
The computation above was independent of $\bx$. So, $\sigma$ allows us to change the scale of all vector norms  simultaneously.

We now prove Theorem \ref{thm:invariant} by proving each item of Definition \ref{def:invariant} for $\EE{} X\tup$.
Denote $\Tilde{\bx}_i= \bx_i - \frac{1}{n}\sum_{j=1}^n \bx_j$ for $1\leq i\leq n$  and $\Tilde{\bx}_{n+1}= \bm \nu_c  - \frac{1}{n}\sum_{j=1}^n \bx_j$.  We now examine the geometry of $\lrc{ {\Tilde{\bx}_i} }_{i=1}^n$. To that end, we choose a coordinate system where $\bx_1=(1,0,...,0)$ and $\bm \nu_c=(0,\frac{1}{n-1},0,...,0) $. We may choose any coordinate system because we calculate the expectation over a Gaussian distribution which is spherically symmetric, and the choice we made is allowed since $\bx_1$ and $\bm\nu_c$ are orthogonal.

Then,
\begin{align}
   \Tilde{\bx}_1 &= \bx_1-  \frac{1}{n}\sum_{i=1}^n \bm x_i \\
   &= \bx_1 -  \frac{1}{n} \bm x_1  -\frac{n-1}{n}\bm \nu_c  \\
   &= \lrp{ \frac{n-1}{n} ,  -\frac{1}{n}, 0, ..., 0  } 
\end{align} 

and similarly,
  
 \begin{align}
   \Tilde{\bx}_{n+1} &= \bm \nu_c -  \frac{1}{n}\sum_{i=1}^n \bm x_i \\
   &= \bm \nu_c  -  \frac{1}{n} \bm x_1  -\frac{n-1}{n}\bm \nu_c  \\
   &= \lrp{ -\frac{1}{n} ,  \frac{1}{n(n-1)}, 0, ..., 0  }. 
\end{align} 

Hence,

$$ \norm{ \Tilde{\bx}_1}^2=   \frac{n^2 -2n +2 }{n^2} ~~~~ \text{and}  ~~~~  \norm{ \Tilde{\bx}_{n+1}}^2=   \frac{n^2 -2n +2 }{n^2\lrp{n-1}^2 } $$

This shows that if we pick $\sigma^2 = \frac{2\alpha}{d} $, we have proved item 1 since

\begin{align}
\EE{} \norm{\bm x\tup_1 }^2 &= \EE{} \R \lrp{W \bx_1 -\bm \mu\tu} \cdot  \R \lrp{W \bx_1 -\bm \mu\tu} \\
   &=  \EE{} \R \lrp{W \Tilde{\bx}_1  } \cdot  \R \lrp{W  \Tilde{\bx}_1} \\
   &=  \frac{d\sigma^2}{2}  \norm{  \Tilde{\bx}_1 }^2 = 1,
\end{align}

and similarly,  item 2 since
\begin{align}
\EE{} \norm{\hat{ \bm \nu}_c  }^2 &= \EE{} \R \lrp{W \bm \nu_c -\bm \mu\tu} \cdot  \R \lrp{W \bm \nu_c -\bm \mu\tu} \\
   &=  \EE{} \R \lrp{W \Tilde{\bx}_{n+1}  } \cdot  \R \lrp{W  \Tilde{\bx}_{n+1}} \\
   &=  \frac{d\sigma^2}{2}  \norm{  \Tilde{\bx}_{n+1} }^2 = \frac{1}{(n-1)^2}.
\end{align}

Item 3 follows from two simple observations: 
\begin{itemize}
    \item $ \phase{ \Tilde{\bx}_1 }{\Tilde{\bx}_{n+1} }=\pi$
    \item If $\bu,\bv$ are vectors such that  $\bu=-a\bv$ for $a>0$, then $\R(\bu)\cdot \R(\bv)=0$.
\end{itemize}

So, by the first observation we have $\Tilde{\bx}_1  = -a \Tilde{\bx}_{n+1} $ for some $a>0$. So,  $W\Tilde{\bx}_1  = -a W\Tilde{\bx}_{n+1} $   holds for any $W$. The second observation then implies  
$$
  \bx\tup_1  \cdot\hat{ \bm \nu}_c =  \R\lrp{ W\Tilde{\bx}_1 }  \cdot  \R\lrp{W\Tilde{\bx}_{n+1}} = 0 ,
    $$
    which concludes the proof of item 3.

Finally, item 4 is trivial since   $\bm x_i =\bm \nu_{c}   $ for all $2 \leq i \leq n$ and  $f(\bx)= \EE{} \R\lrp{W \bx  -\bm \mu\tu}$ is a function of $\bx$.

The proof of items 1, 2, and 3 of Theorem~\ref{thm:rand_stb} follows the same lines of the proof for Theorem~\ref{thm:invariant}, this time with $\alpha=\frac{n^2}{(n-1)^2R^2}$.

For item 4, we require a more intricate expectation calculation than in equation~\ref{eq:norm}. We need  the expected  inner product between $\R(\rw   \bx)$ and $\R(\rw \by)$. By equation 6 in~\cite{cho} we have:
\begin{align}\label{eq:dual_relu}
    \EE{} \R \lrp{W  \bx } \cdot \R \lrp{ W  \by } = \frac{d\sigma^2 \norm{x} \norm{y} }{2}    \frac{\sqrt{1-\rho^2} +\left( \pi -\cos^{-1}(\rho) \right)\rho}{\pi} \coloneqq K(\bx,\by)
\end{align}

 where $\rho\coloneqq \frac{\bx\cdot \by}{\norm{\bx} \norm{\by}}$ is the similarity between vectors $\bx$ and $\by$.

 The function appearing in equation~\ref{eq:dual_relu}  has an important property:

 $$ K(\bx,\by) > \bx \cdot \by   $$

for $\sigma^2 =\frac{2}{d}$ and $\bx\neq \by$. 

For $\bx_i\neq \bx_j$, this property  implies that for $\sigma^2 = \frac{2}{d}$ we have:

\begin{align}
  \norm{  \bx_i - \bx_j   }^2 &= \norm{  \tilde{\bx}_i - \tilde{\bx}_j  }^2  \\
&= \norm{ \tilde{\bx}_i }^2 +   \norm{ \tilde{\bx}_j }^2 -2 \tilde{\bx}_i \cdot \tilde{\bx}_j \\
&= \EE{} \norm{ \R \lrp{W\bx_i  - \bm \mu\tu   }}^2 + \norm{\R  \lrp{ W\bx_j  - \bm \mu\tu } }^2  -2 \tilde{\bx}_i \cdot \tilde{\bx}_j \\
&>  \EE{} \norm{ \R \lrp{W\bx_i  - \bm \mu\tu   }}^2 + \norm{\R  \lrp{ W\bx_j  - \bm \mu\tu } }^2 - 2 K( \tilde{\bx}_i ,\tilde{\bx}_j) \\
&=      \EE{} \norm{ \R \lrp{W\bx_i  - \bm \mu\tu   }-   \R \lrp{W\bx_j   - \bm \mu\tu   } }^2  \\
&= \EE{} \norm{ \bx\tup_i - \bx\tup_j     } ^2 
\end{align}

The result also holds if we take $\sigma^2 = \frac{2\alpha}{d}$ with $\alpha = \frac{n^2}{(n-1)^2R^2}$. This is because ReLU is homogeneous $\R(\alpha x)=\alpha \R(x)$ for $\alpha > 0$, so that we have 
$$ \EE{} \norm{ \bx\tup_i - \bx\tup_j}^2 < \alpha \norm{  \bx_i - \bx_j   }^2 < \norm{  \bx_i - \bx_j   }^2$$
since $\alpha < 1$. This concludes the proof of item 4.

\end{document}